\documentclass{article}

\PassOptionsToPackage{numbers,sort&compress}{natbib}
\usepackage[main,final]{neurips_2026}
\renewcommand{\acksection}{%
  \section*{Acknowledgments and disclosure of funding}%
}
\usepackage{amsmath,amsfonts,bm}

\def\eqref#1{equation~\ref{#1}}
\def\1{\bm{1}}

\def\vc{{\bm{c}}}

\def\vh{{\bm{h}}}

\def\vp{{\bm{p}}}
\def\vq{{\bm{q}}}

\def\vt{{\bm{t}}}

\def\vv{{\bm{v}}}

\def\vx{{\bm{x}}}

\def\vz{{\bm{z}}}

\def\mI{{\bm{I}}}

\def\mQ{{\bm{Q}}}

\DeclareMathAlphabet{\mathsfit}{\encodingdefault}{\sfdefault}{m}{sl}
\SetMathAlphabet{\mathsfit}{bold}{\encodingdefault}{\sfdefault}{bx}{n}

\newcommand{\E}{\mathbb{E}}

\usepackage[utf8]{inputenc} %
\usepackage[T1]{fontenc}    %
\usepackage{url}            %
\usepackage{booktabs}       %
\usepackage{amsfonts}       %
\usepackage{nicefrac}       %
\usepackage{microtype}      %
\usepackage{xcolor}         %

\usepackage{amsmath, array}%
\usepackage{amssymb}
\usepackage{bm}
\usepackage{textcomp}
\usepackage{siunitx}
\usepackage{tikz}
\usepackage{graphicx}
\usetikzlibrary{positioning, shapes.geometric, arrows}
\usetikzlibrary{decorations.pathreplacing}
\usetikzlibrary{calc}
\usepackage{comment}
\usepackage{subcaption}

\usepackage{etoc}
\usepackage[colorlinks=true]{hyperref}
\newcommand{\doi}[1]{doi: \begingroup\urlstyle{rm}%
  \href{https://doi.org/#1}{\nolinkurl{#1}}\endgroup}
\usepackage[noabbrev]{cleveref}
\crefname{appsec}{appendix}{appendices}
\Crefname{appsec}{Appendix}{Appendices}

\usepackage{orcidlink}
\usepackage{threeparttable}  
\usepackage{multirow}
\usepackage[ruled,vlined]{algorithm2e}
\usepackage{capt-of}

\graphicspath{{figs/}}
\title{FlowBatt: Flow Matching for Probabilistic Battery Degradation Prediction}

\author{%
  Hamidreza Eivazi$^{1,}$\thanks{Corresponding author: \texttt{h.eivazi@tu-braunschweig.de}~(https://orcid.org/0000-0003-3650-4107)} \quad
  Eslam Sharaawy$^2$ \quad
  Stefan Wittek$^2$ \\
  \textbf{André Hebenbrock}$^{3}$ \quad
  \textbf{Raphael Ginster}$^{4}$ \quad
  \textbf{Steffen Blömeke}$^{5}$\\ \\
  $^1$\,Division Data-Driven Modeling of Mechanical Systems, Institute of Applied Mechanics, \\Technische Universität Braunschweig, Germany\\[2pt]
  $^2$\,Institute for Software and Systems Engineering, Clausthal University of Technology, Germany \\[2pt] 
  $^3$\,Institute of Chemical and Electrochemical Process Engineering, \\Clausthal University of Technology, Germany\\[2pt] 
  $^4$\,Institute of Automotive Management and Industrial Production, \\Technische Universität Braunschweig, Germany\\[2pt] 
  $^5$\,Institute of Machine Tools and Production Technology, \\Technische Universität Braunschweig, Germany
}

\begin{document}
\maketitle

\begin{abstract}
Battery degradation remains a central challenge in the development and broad application of sustainable energy technologies. Accurate degradation prediction is challenging, as battery aging emerges from complex and heterogeneous interactions among cycling behavior, operating conditions, and cell chemistry. Existing machine-learning approaches typically focus on deterministic end-of-life predictions or modeling degradation curves within fixed chemistry settings, leaving probabilistic state of health (SOH) trajectory modeling across heterogeneous chemistries underexplored. We introduce FlowBatt, a use-inspired conditional generative framework that adapts flow matching with a diffusion transformer (DiT) backbone to battery degradation trajectory modeling. FlowBatt models full SOH trajectories as probabilistic degradation processes conditioned on early-cycle capacity data. To strengthen this conditioning pathway, we use explainable AI analysis to diagnose and improve the encoder that maps early-cycle capacity matrices into conditioning vectors. We evaluate FlowBatt on five public benchmark settings spanning diverse chemistries and aging conditions, comparing it with supervised and diffusion-based trajectory models as well as established baselines for remaining useful life (RUL) prediction under shared data splits. FlowBatt achieves the best trajectory-wise SOH prediction performance on all five datasets, with errors below 2.5\%, and delivers competitive RUL prediction performance. By generating multiple plausible trajectories, FlowBatt provides empirical prediction intervals for battery degradation modeling. These results indicate that flow-matching-based trajectory modeling is a promising framework for probabilistic battery health prediction.
\end{abstract}

\section{Introduction}

\textbf{Lithium-ion batteries and degradation.}
Lithium-ion (Li-ion) batteries are central to modern energy-storage applications, including portable electronics, electric vehicles, and stationary storage systems \citep{Blomgren2017TheBatteries}. Their widespread use is driven by favorable properties such as high energy density, long cycle life, and low self-discharge \citep{Galeotti2015PerformanceSpectroscopy,Vetter2005AgeingBatteries}. Despite these advantages, battery degradation remains a major limitation for reliability, safety, and economic viability \citep{Chombo2020ABattery,Blomeke2022,Ginster2024CircularImpacts}. A key manifestation of this degradation is capacity fade, which arises from complex electrochemical and mechanical mechanisms influenced by operating conditions, usage history, and cell chemistry \citep{Broussely2005MainBatteries,Edge2021LithiumKnow,OKane2022Lithium-ionIt}. This complexity makes accurate degradation forecasting challenging.

\textbf{Machine learning for battery health prediction.}
Battery degradation is commonly characterized through state of health (SOH), which measures the remaining capacity relative to the nominal capacity, and remaining useful life (RUL), which estimates the cycles or time remaining before end of life (EOL) \citep{Li2022TowardsReview,Cui2022ANetwork,Ren2023ABatteries,Li2023TheBatteries}. Since both quantities depend on complex and nonlinear degradation dynamics, machine learning has become an increasingly important tool for battery health prediction \citep{Rauf2022MachineModelling}. Prior work has explored a wide range of methods, including linear models, support vector machines, Gaussian processes, convolutional and recurrent neural networks, and transformers \citep{Li2022TowardsReview,Li2023TheBatteries,Rauf2022MachineModelling,Ren2023ABatteries,Wang2021,Hu2026BatteryGPT}. Most of these approaches are formulated as supervised regression models and focus on endpoint prediction, such as a scalar SOH value at a specific cycle or the RUL of a cell. 
Standardized benchmarks such as BatteryML \citep{Zhang2023BatteryML:AnDegradation} and BatLiNet \citep{Zhang2025} have improved comparability across datasets and models. However, existing benchmarks also highlight an important limitation of the current literature: most methods are optimized for deterministic point prediction, especially RUL, while probabilistic modeling of full SOH degradation trajectories remains much less explored. A notable recent exception is the diffusion-based work of \citet{Luo2023State-of-HealthLearning}, which points to the potential of generative modeling for battery SOH prediction.

\textbf{Contribution to literature.}
This work advances the domain of battery degradation prediction in several aspects. 
First, we introduce \emph{FlowBatt}, a use-inspired conditional generative framework that adapts flow matching to probabilistic battery degradation modeling. Rather than proposing a new flow-matching objective, our contribution is to formulate battery health prediction as conditional full-trajectory SOH generation from early-cycle capacity data. Leveraging a diffusion transformer (DiT) backbone, FlowBatt captures the stochastic and nonlinear dynamics of battery aging as a probabilistic, trajectory-level process more effectively than traditional supervised approaches (\cref{sec:methodology}).
Second, we use explainable AI to analyze the encoder that maps early-cycle capacity matrices to conditioning vectors, identify failure modes in a ReLU-based encoder, and adopt a more robust LeakyReLU encoder for proposed models (\cref{app:cnn}).
Third, we demonstrate the benefits of flow-based generative modeling by benchmarking FlowBatt against diffusion processes and supervised learning methods. Our results show that FlowBatt achieves the best performance in SOH prediction across diverse aging conditions and chemistry settings, while also delivering competitive performance compared to the state-of-the-art in RUL prediction (\cref{sec:results}). 
Finally, we establish FlowBatt as a probabilistic model for battery degradation, capable of unifying SOH trajectory forecasting, uncertainty quantification, and degradation-curve synthesis within a single framework, enabling future extensions and adaptations across diverse energy storage applications (\cref{sec:conclusions}). The source code and pre-trained models are publicly available at \url{https://github.com/HamidrezaEiv/FlowBatt}.

\section{Methodology}
\label{sec:methodology}

Denoising diffusion probabilistic models (DDPMs) \citep{ddpm} are a flexible family of generative models that learn to approximate a data distribution through a gradual noising-denoising process. The forward process is defined as a parameterized Markov chain that progressively corrupts the data with Gaussian noise until the signal is indistinguishable from random noise. The reverse process, also known as the denoising or generative process, is learned by the model and gradually removes noise step by step to reconstruct samples resembling the training data. Training is performed using a variational objective that optimizes a lower bound on the data likelihood across many diffusion steps. Once trained, DDPMs can generate diverse, high-quality samples, which has led to widespread applications in image and audio synthesis \citep{dhariwal2021diffusion,Ho2022,yang2024surveydiffusionmodelstime}, data augmentation \citep{luzi2024boomeranglocalsamplingimage}, and scientific modeling \citep{bastek2024physicsinformeddiffusionmodels,Frrutter2024,Li2024,Zhang2024}. 

Flow matching \citep{lipman2023,LiuG023,AlbergoV23} is an alternative to diffusion training that directly learns continuous-time generative dynamics. Instead of discretizing the forward and reverse processes into thousands of noisy steps as in DDPMs, flow matching estimates an ordinary differential equation (ODE) that smoothly transports a base distribution (e.g., Gaussian noise) into the target data distribution. This approach avoids the stochasticity and step-by-step denoising of DDPMs and enables more efficient training with fewer discretization steps. Moreover, flow matching naturally admits exact likelihood estimation, making it both expressive and computationally attractive for high-dimensional scientific data. In this work, we adopt flow matching to train our generative model of battery degradation.  

Transformers have become a powerful backbone for diffusion-based generative models, giving rise to diffusion transformers (DiTs) \citep{PeeblesDIT}. Unlike convolutional UNet backbones, DiTs leverage global self-attention to capture long-range dependencies in the data. Diffusion transformers typically operate on tokenized patches of the input, and use an effective conditioning mechanism based on adaptive layer normalization \citep{PerezSVDC18} for conditioning on auxiliary information such as class labels or text encodings. In this work, the auxiliary information corresponds to the battery's performance during the early cycle life. Our approach employs DiTs directly on raw data representations of battery degradation. This design combines the scalability and expressivity of transformers with the probabilistic training of flow matching to capture the nonlinear, stochastic dynamics of battery aging. 

\subsection{\texorpdfstring{Diffusion and flow-matching foundations}{Diffusion and flow-matching foundations}}

This subsection introduces the diffusion and flow-matching objectives used by DiffBatt and FlowBatt, respectively.
We represent a cycle-indexed SOH trajectory at generative time $t$ as $\vx_t=(x_t^{(1)},\ldots,x_t^{(T)})\in\mathbb{R}^T$, where $T$ is the number of cycles, $t$ denotes generative time, and $i$ denotes the sampled battery-cycle index. In diffusion, $\vx_t$ is the noisy trajectory, whereas in flow matching it is the time-$t$ interpolant.

Denoising diffusion models (DDMs) learn to transform a simple prior distribution, typically a unit Gaussian $\mathcal{N}(\bm{0}, \mI)$, into an unknown data distribution $q(\vx)$ by gradually denoising samples corrupted with Gaussian noise \citep{ddpm}. A fixed \textit{forward process} incrementally adds noise to data $\vx_0 \sim q(\vx)$ according to a variance schedule, and a neural network is trained to approximate the \textit{reverse process} that removes noise step by step. With the standard parameterization suggested in \cite{ddpm}, the training objective reduces to a denoising score matching loss, where the network $\bm{\epsilon}_\theta(\vx_t, t)$ predicts the injected Gaussian noise
\begin{equation}
  \mathcal{L}_{\mathrm{simple}}(\theta) := \E_{t, \vx_0, \bm{\epsilon}}\left[\|\bm{\epsilon} - \bm{\epsilon}_\theta(\vx_t, t)\|^2\right].
\end{equation}
Here, $\bm{\epsilon}\sim\mathcal{N}(\bm{0},\mI)$ denotes the sampled Gaussian noise used to construct $\vx_t$ in the forward process.
For conditional modeling, auxiliary information $\vc$ can be incorporated into the reverse process by conditioning the network $\bm{\epsilon}_\theta(\vx_t, t, \vc)$, often enhanced with classifier-free guidance \citep{diffusionguidance}. While diffusion models achieve strong generative performance, sampling requires simulating many reverse steps, which motivates alternative formulations. 

Instead of constructing a discrete forward diffusion process, flow matching \citep{lipman2023, LiuG023} formulates generative modeling as learning a continuous-time transport map between a tractable prior distribution $p(\vx)$, typically $\mathcal{N}(\bm{0}, \mI)$, and the unknown data distribution $q(\vx)$. The approach defines a family of interpolant distributions $\{p_t(\vx)\}_{t \in [0,1]}$ that smoothly connects the prior $p_0(\vx)=p(\vx)$ to the target $p_1(\vx)=q(\vx)$. This interpolation is governed by an ordinary differential equation (ODE) of the form  
\begin{equation}
    \frac{d\vx_t}{dt} = \vv_\theta(\vx_t, t), \quad \vx_0 \sim p(\vx),
\end{equation}
where $\vv_\theta$ is a neural network parameterized vector field trained to approximate the unknown probability flow that transports samples from prior to data. To obtain a tractable training objective, one introduces a \emph{target vector field} defined by the time derivative of a prescribed stochastic interpolant $\vx_t = \alpha(t)\vx_0 + \beta(t)\vx_1$ with $(\vx_0, \vx_1) \sim p(\vx) \times q(\vx)$. For example, choosing linear interpolation $\alpha(t)=1-t$, $\beta(t)=t$ yields  
\begin{equation}
    \tilde{\vv}(\vx_t, t \vert \vx_0, \vx_1) = \frac{d}{dt} \left[ \alpha(t)\vx_0 + \beta(t)\vx_1 \right] = \dot{\alpha}(t)\vx_0 + \dot{\beta}(t)\vx_1.
\end{equation}
The model $\vv_\theta(\vx_t, t)$ is then trained to match this target vector field in expectation:  
\begin{equation}
    \mathcal{L}_{\text{FM}}(\theta) := \E_{t, \vx_0 \sim p(\vx), \vx_1 \sim q(\vx)}\Big[\|\vv_\theta(\vx_t, t) - \tilde{\vv}(\vx_t, t \vert \vx_0, \vx_1)\|^2\Big].
\end{equation}
This training can be interpreted as regressing the dynamics of sample trajectories along linear interpolation paths from the prior to the data distribution. Once trained, the model generates samples by integrating the learned ODE  
\begin{equation}
    \vx_1 = \vx_0 + \int_{0}^{1} \vv_\theta(\vx_t, t)\, dt, \quad \vx_0 \sim p(\vx),
\end{equation}
which requires significantly fewer steps than simulating the reverse diffusion process. Moreover, flow matching provides a unifying framework: with appropriate choices of interpolants, it can recover score-based diffusion models as a special case while also supporting more efficient sampling schemes. For conditional generation, the conditioning variable $\vc$ is incorporated directly into the learned vector field $\vv_\theta(\vx_t, t, \vc)$. 

\subsection{Architecture}
\label{subsec:architecture}

\begin{figure}[h]
  \centering
  \input{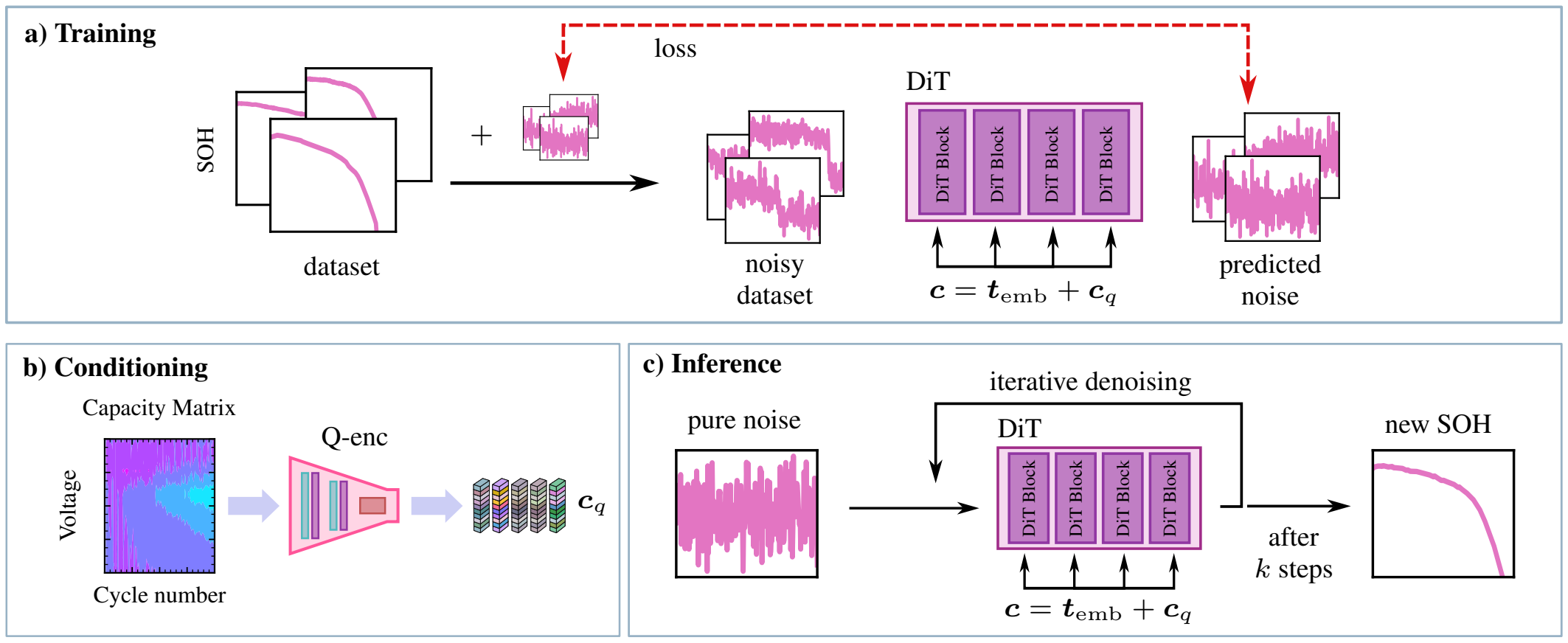}
  \caption{\small Schematic view of the model architecture. Adapted and modified from \cite{Frrutter2024}. Modifications include context-specific changes.}
  \label{fig:architecture}
\end{figure}

\Cref{fig:architecture} provides an overview of the FlowBatt architecture and its workflow. The top panel depicts the diffusion-based training stage, where noisy SOH trajectories are progressively denoised through a stack of DiT blocks conditioned on early-cycle capacity matrices. The bottom panels illustrate the auxiliary pathway and the generation stage, where the capacity matrix is first encoded by a CNN into a compact embedding, which is combined with the timestep embedding to modulate the DiT backbone and generate complete SOH trajectories.

\textbf{DiT backbone and block design.}
FlowBatt instantiates a compact Diffusion Transformer (DiT) for long one-dimensional sequences representing SOH during the battery cycle life. Given the noisy, cycle-indexed SOH trajectory $\vx_t\in\mathbb{R}^{T}$, we first lift it to a token sequence via a $1{\times}1$ convolution giving $\vh_0\in\mathbb{R}^{T\times d}$, where $d$ is the hidden width (number of channels). After preprocessing, $T=256$, with one SOH value sampled every ten physical cycles. We add fixed one-dimensional sinusoidal positional embeddings $\vp\in\mathbb{R}^{T\times d}$ to preserve temporal order. The core of the network is a stack of $N$ identical DiT blocks, each following the Transformer paradigm of a pre-normalized residual connection with a multi-head self-attention (MSA) sublayer, followed by a feed-forward multilayer perceptron (MLP) sublayer. The stack is followed by a final projection layer that applies an adaptive normalization conditioned on the conditioning vector and then maps the hidden states back to the data domain. This ensures that conditioning information influences not only the internal block dynamics but also the final predicted vector field used in flow matching.

\textbf{Conditioning mechanism.}
In the adaptive layer norm (adaLN)-Zero formulation \citep{PeeblesDIT}, each MSA and MLP sublayer is modulated by per-channel scale-offset and shift parameters $(\gamma, \beta)$ applied to the normalized activations, together with residual scaling factors $\alpha$ applied directly before the residual addition. All modulation parameters are produced from a single global conditioning vector $\vt_{\mathrm{emb}} + \vc_{\vq}$ that merges a timestep embedding $\vt_{\mathrm{emb}} = \phi(t)$ and an auxiliary embedding $\vc_{\vq}$ obtained by encoding the capacity matrix corresponding to the early cycle life of the battery. The conditioning vector is passed through a small multilayer perceptron that produces six $d$-dimensional vectors as per-channel modulation parameters. The residual scalars $\alpha$ are initialized to zero, ensuring that every block starts as an identity mapping and that conditioning effects are introduced gradually during training. This approach has been shown to improve optimization stability and sample quality in large-scale generative modeling \citep{lipman2023}. The detailed computational steps of one DiT block with adaLN-Zero conditioning are provided in \cref{app:ditblock}.

\textbf{CNN encoder.}
\label{par:cnn_encoder}
We employ the concept of \textit{capacity matrix} ($\mQ$) introduced by \citet{attia2021statistical}, as auxiliary information for conditional generation. The capacity matrix is a cycle-by-voltage array of baseline-subtracted discharge-capacity curves. Its construction and rationale are detailed in \cref{app:capacity_matrix}. We utilize the capacity matrix obtained from the early cycle life of the battery, i.e., the first 20 or 100 cycles similar to \cite{attia2021statistical,Severson2019Data-drivenDegradation,Zhang2023BatteryML:AnDegradation}. This choice is driven by the high costs, time, and effort associated with long-term battery testing. Our goal is to leverage early life performance data to predict battery degradation and minimize resource expenditure. To encode $\mQ$ into an embedding ($\vc_q$), we utilize a CNN encoder (see \cref{fig:architecture}\textcolor{red}{b}). We adapt a CNN architecture inspired by \citet{attia2021statistical,Zhang2025}, which has been used for RUL prediction. This model has been reported to lack robustness with respect to different initializations, however, it provides very accurate predictions in some instances. To enhance the robustness of the model we utilize explainable AI methods, specifically Grad-CAM (gradient-weighted class activation mapping) method \citep{Selvaraju2017Grad-CAM:VisualExplanations} to better understand how these deep-learning models make predictions for the RUL task. Further, we identify and eliminate potential modes of failure. Results for this analysis are reported in \cref{app:cnn}. This explainable AI analysis reveals initialization-sensitive failure modes in a ReLU-based CNN encoder, including cases with inactive gradients. We therefore replace ReLU activations with leaky rectified linear unit (LeakyReLU) activations to obtain a more robust XAI-guided CNN encoder. The final architecture consists of two convolutional layers with LeakyReLU activations, each followed by average pooling to reduce spatial dimensions. The resulting feature map is projected with a final convolution, activated, and flattened into a $d$-dimensional vector $\vc_q$. This CNN encoder effectively captures the relevant features from the capacity matrix, enabling FlowBatt to generate accurate and context-aware SOH trajectories.

\begin{figure}[h]
  \centering
  \includegraphics[width=\textwidth]{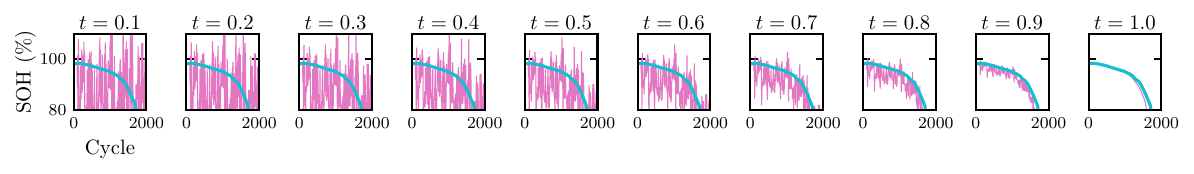}
  \includegraphics[width=\textwidth]{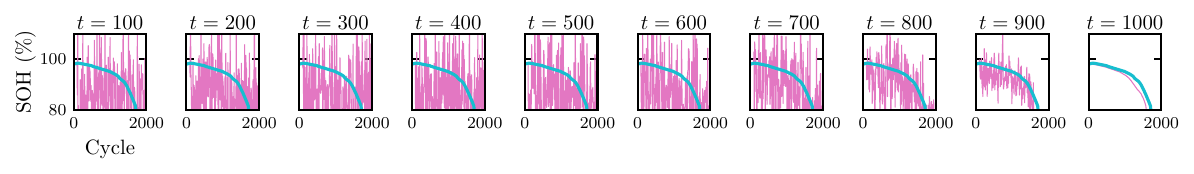}
  \caption{\small Comparison of denoising trajectories for a test sample from the \texttt{MATR-1} dataset using flow matching (top) and the diffusion process (bottom). The pink line shows the generated sample and the cyan line shows the reference for a given capacity matrix. The displayed FlowBatt states correspond to requested ODE output times for visualization.}
  \label{fig:record}
\end{figure}

In summary, our generative models trained via flow matching and diffusion processes combine a DiT backbone with adaLN-Zero conditioning and a CNN-based encoder for auxiliary early-life data, yielding a scalable generative framework that provides empirical prediction intervals. By integrating timestep information with early-cycle capacity matrices, the model learns to generate realistic SOH trajectories that reflect both temporal dynamics and usage context. From these SOH trajectories, metrics such as RUL can be directly computed by applying practical thresholds (e.g., 80\% of nominal capacity). This design positions FlowBatt as a flexible model for battery health prediction, capable of both probabilistic modeling of aging behavior and data-driven synthesis. \Cref{fig:record} illustrates the denoising trajectories of a test sample from the \texttt{MATR-1} dataset using flow matching and diffusion processes. It can be observed that the flow matching approach achieves faster denoising compared to the diffusion process. Note that diffusion requires 1000 reverse steps, while in flow matching we solve the ODE over [0,1].

\subsection{Data}

We conduct a comprehensive evaluation based on several publicly available datasets curated in BatteryML \citep{Zhang2023BatteryML:AnDegradation}. Specifically, we include \texttt{CALCE} \citep{CALCE2,CALCE1}, \texttt{HUST} \citep{HUST}, \texttt{MATR} \citep{CLO,Severson2019Data-drivenDegradation}, \texttt{RWTH} \citep{RWTH}, \texttt{SNL} \citep{SNL}, and \texttt{UL\_PUR} \citep{UL_PUR,UL_PUR2}, covering a range of chemistries including lithium iron phosphate (LFP), lithium cobalt oxide (LCO), nickel manganese cobalt oxide (NMC), nickel cobalt aluminum oxide (NCA), and a combination of NMC and LCO (NMC\_LCO). These datasets span diverse aging conditions, including differences in capacities, voltages, ambient temperatures, state of charge (SOC), and degradation characteristics, and therefore provide a broad spectrum of battery aging behaviors. 

In this study, we follow the benchmark protocol of BatLiNet \citep{Zhang2025}, which derives five evaluation datasets under different aging conditions. Specifically, BatLiNet defines two \texttt{MATR} evaluation sets (\texttt{MATR-1} and \texttt{MATR-2}) to maintain comparability with existing models. In addition, the \texttt{HUST} dataset, which uses LFP cells similar to \texttt{MATR}, is included to evaluate model adaptation to different cycling protocols. Furthermore, BatLiNet constructs a \texttt{MIX} dataset by aggregating all remaining public datasets, capturing diverse aging conditions such as different ambient temperatures, packaging structures, and cathode chemistries (NMC, LCO, and NCA). From the \texttt{MIX} dataset, two benchmark settings are defined: \texttt{MIX-100}, which evaluates the prediction of the 80\% EOL point using only the first 100 cycles, and \texttt{MIX-20}, a more challenging setup requiring the prediction of the number of cycles until 90\% capacity using only the first 20 cycles. 

While these benchmark definitions were originally introduced for RUL prediction, in this work we use the same datasets and data splits for both SOH trajectory prediction and RUL evaluation. This allows us to assess full degradation-trajectory forecasting under the same standardized experimental conditions and to retain direct comparability with prior benchmark studies. As in \cite{Zhang2025}, batteries that reached EOL prematurely during the early cycles are excluded from the analysis. Further details on datasets and preprocessing are provided in \cref{app:data} and \cite{Zhang2025,Zhang2023BatteryML:AnDegradation}.

\section{Results and discussion}
\label{sec:results}

Our generative models can be utilized for trajectory-wise SOH prediction, RUL prediction, and SOH synthesis. Here, SOH synthesis refers to the generation of new SOH curves for data augmentation. Among these tasks, trajectory-wise SOH prediction is the primary focus of this work, since FlowBatt is designed to model full degradation processes rather than only scalar quantities. Cycle-wise SOH prediction estimates SOH at a predefined cycle, whereas RUL prediction estimates the first cycle at which SOH reaches or falls below the EOL threshold. Our trajectory models provide both scalar predictions without task-specific training. SOH and RUL prediction are closely related tasks in battery health management: SOH describes the temporal evolution of battery health, whereas RUL can be derived as a downstream EOL indicator from the predicted SOH trajectory.

In this section, we conduct experiments based on the BatLiNet benchmark tests \citep{Zhang2025} and compare the results across models. We first evaluate trajectory-wise SOH prediction to assess the ability of the models to forecast full degradation trajectories, which is the most relevant setting for FlowBatt. For completeness and comparability with BatteryML-style evaluation, we also evaluate cycle-wise SOH prediction. We then examine RUL prediction derived from the predicted SOH trajectories under the same benchmark setting. We report results for three models trained using supervised learning, diffusion processes, and flow matching, referred to as Transformer, DiffBatt, and FlowBatt, respectively.

In industrial applications, trajectory-wise SOH prediction supports continuous health monitoring, degradation-aware operation, warranty assessment, and long-term maintenance planning. Cycle-wise SOH prediction supports health assessment at a particular inspection or operating cycle, while RUL provides an actionable endpoint for maintenance scheduling and battery replacement. Evaluation metrics, EOL thresholds, and additional results are provided in \cref{app:soh_results_additional}.

\subsection{SOH prediction}
\label{sec:soh_prediction}

In practical applications, estimating a battery's SOH requires assessing the current discharge capacity under standardized conditions using reference performance tests and historical cycling data. However, the discrepancy between real-world battery usage and these standardized conditions poses significant challenges in obtaining precise ground-truth labels, which complicates accurate SOH prediction. Consequently, establishing robust and reproducible benchmark settings for SOH prediction remains an ongoing effort within the research community.
In this work, we primarily evaluate SOH prediction in a \emph{trajectory-wise} setting. In this setting, the models predict the full SOH trajectory, and the RMSE is computed over the degradation curve up to the EOL point, padding the remainder of the sequence with the final observed value. Results are then reported as the mean RMSE across all test samples. Since SOH is expressed as a percentage of the nominal capacity, the RMSE is also reported in percentage units. Further details on the experimental setup and error metrics are provided in \cref{app:soh_results_additional}.

Most established RUL baselines predict a scalar and require adaptation to produce full SOH trajectories. We include MLP as a standard neural baseline and RF for its best reported RUL performance on BatteryML's largest dataset, MIX \citep{Zhang2023BatteryML:AnDegradation}, together with BatLiNet \citep{Zhang2025}, a state-of-the-art RUL predictor. These three baselines are adapted for trajectory-wise SOH prediction. For RUL, we retain the broader set of direct scalar predictors.

\begin{table}[h]
  \centering
  \caption{\small  Trajectory-wise SOH prediction results reported as mean RMSE (\%) over the predicted degradation trajectories. The $\pm$ values denote standard deviations of RMSE across ten generative draws from one trained checkpoint for FlowBatt and DiffBatt, and across ten training initializations for the other models. Best results are shown in bold and second-best results are underlined. This table evaluates full-trajectory SOH prediction. RMSE is computed over the predicted degradation trajectories up to EOL and therefore measures a different task from the cycle-wise SOH setting reported in \cref{app:soh_cyclewise}.}
  \label{tab:soh-trajectorywise}
  \footnotesize
  \begin{tabular}{@{}lccccc@{}}
  \toprule
  Method & \texttt{MATR-1} & \texttt{MATR-2} & \texttt{HUST} & \texttt{MIX-20} & \texttt{MIX-100} \\
  \midrule
  BatLiNet & 1.86 $\pm$ 0.06 & 3.62 $\pm$ 0.24 & 2.87 $\pm$ 0.23 & 2.04 $\pm$ 0.06 & 2.34 $\pm$ 0.22 \\
  RF (BatteryML) & 2.01 $\pm$ 0.04 & 2.50 $\pm$ 0.01 & 3.57 $\pm$ 0.03 & 3.42 $\pm$ 0.02 & 5.45 $\pm$ 0.04 \\
  MLP (BatteryML) & 1.97 $\pm$ 0.11 & 3.06 $\pm$ 0.17 & 4.11 $\pm$ 0.13 & 2.23 $\pm$ 0.09 & 3.13 $\pm$ 0.14 \\
  Transformer (ours) & 1.09 $\pm$ 0.07 & 2.25 $\pm$ 0.22 & 3.18 $\pm$ 0.16 & 1.42 $\pm$ 0.10 & 1.76 $\pm$ 0.06 \\
  DiffBatt (ours) & \underline{0.92 $\pm$ 0.04} & \underline{1.85 $\pm$ 0.03} & \underline{2.44 $\pm$ 0.14} & \underline{1.15 $\pm$ 0.08} & \underline{1.71 $\pm$ 0.09} \\
  FlowBatt (ours) & \textbf{0.87 $\pm$ 0.02} & \textbf{1.77 $\pm$ 0.04} & \textbf{2.10 $\pm$ 0.12} & \textbf{0.88 $\pm$ 0.03} & \textbf{1.41 $\pm$ 0.02} \\
  \bottomrule
  \end{tabular}
\end{table}
The trajectory-wise SOH results are reported in \cref{tab:soh-trajectorywise}. This setting directly evaluates the models on their ability to forecast full degradation trajectories and is therefore the most representative evaluation for FlowBatt. The supervised Transformer baseline provides a useful point of comparison, achieving reasonable performance on some datasets such as \texttt{MATR-1} (1.09$\pm$0.07) and \texttt{MIX-20} (1.42$\pm$0.10), but showing noticeably higher errors on more challenging datasets such as \texttt{HUST} (3.18$\pm$0.16). DiffBatt improves upon the Transformer on all five datasets and achieves the second-best results across the benchmark, indicating that diffusion-based generative modeling already provides a strong improvement over purely supervised trajectory prediction.
FlowBatt achieves the lowest mean trajectory-wise SOH RMSE on all five datasets. In particular, it reaches an RMSE of 0.87$\pm$0.02 on \texttt{MATR-1}, 1.77$\pm$0.04 on \texttt{MATR-2}, 2.10$\pm$0.12 on \texttt{HUST}, and 0.88$\pm$0.03 on the heterogeneous \texttt{MIX-20} dataset. These results indicate that flow matching provides a reliable and scalable approach for SOH trajectory forecasting across diverse battery chemistries and operating conditions. FlowBatt also provides a unified generative framework for full-trajectory prediction, uncertainty quantification, and synthesis. Overall, the trajectory-wise results confirm that FlowBatt is particularly strong when the task requires modeling the temporal evolution of degradation rather than only isolated SOH targets.

\subsection{Uncertainty quantification}
\label{sec:soh_uncertainty}

\begin{figure}[htb]
  \centering
  \includegraphics[width=\textwidth]{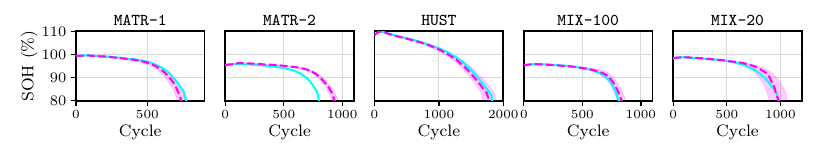}
  \caption{\small Example SOH prediction intervals obtained from FlowBatt across the five benchmark datasets. The cyan curve denotes the reference SOH trajectory, the magenta dashed curve denotes the median prediction, and the shaded region denotes the empirical 90\% prediction interval. The displayed batteries are chosen as the test sample with median predictive uncertainty.}
  \label{fig:soh-interval-examples}
\end{figure}

As FlowBatt is a probabilistic generative model, it can generate multiple plausible SOH trajectories for the same input capacity matrix. We therefore evaluate its uncertainty estimates by constructing empirical 90\% prediction intervals from independently generated trajectories and measuring the fraction of reference SOH values contained within these intervals. The average interval width is also reported to assess the sharpness of the uncertainty bands. The interval definitions and comparisons with DiffBatt, Quantile RF, and Quantile MLP are provided in \cref{app:uq_metrics}.
\begin{table}[h]
      \centering
  \caption{\small  Empirical SOH prediction-interval coverage of FlowBatt for a nominal 90\% interval. Coverage is the fraction of reference SOH values contained within the generated prediction interval. SOH width denotes the average interval width in percentage points. Best coverage values are shown in bold.}
  \label{tab:soh-uncertainty-coverage}
  \footnotesize
  \setlength{\tabcolsep}{3.5pt}
  \begin{tabular}{@{}llccccc@{}}
  \toprule
  Samples & Metric & \texttt{MATR-1} & \texttt{MATR-2} & \texttt{HUST} & \texttt{MIX-20} & \texttt{MIX-100} \\
  \midrule
  \multirow{2}{*}{10}
  & SOH coverage & 86.0\% & 67.0\% & 65.0\% & 83.6\% & 81.7\% \\
  & SOH width & 0.30\% & 0.42\% & 1.34\% & 0.52\% & 0.64\% \\
  \midrule
  \multirow{2}{*}{50}
  & SOH coverage & 87.4\% & 70.0\% & 71.0\% & 86.4\% & 85.4\% \\
  & SOH width & 0.36\% & 0.50\% & 1.62\% & 0.63\% & 0.77\% \\
  \midrule
  \multirow{2}{*}{100}
  & SOH coverage & 88.0\% & 70.0\% & \textbf{73.0\%} & 87.0\% & 86.0\% \\
  & SOH width & 0.37\% & 0.51\% & 1.65\% & 0.65\% & 0.80\% \\
  \midrule
  \multirow{2}{*}{500}
  & SOH coverage & \textbf{88.4\%} & \textbf{70.2\%} & 72.8\% & \textbf{87.4\%} & \textbf{86.3\%} \\
  & SOH width & 0.38\% & 0.52\% & 1.65\% & 0.67\% & 0.83\% \\
  \bottomrule
  \end{tabular}
\end{table}
\Cref{tab:soh-uncertainty-coverage} shows that FlowBatt provides informative SOH prediction intervals, and \cref{fig:soh-interval-examples} illustrates representative examples across the evaluated datasets. With 100 generated samples, the empirical coverage reaches 88.0\% on \texttt{MATR-1}, 87.0\% on \texttt{MIX-20}, and 86.0\% on \texttt{MIX-100}, while keeping the average interval width below one percentage point on these datasets. Increasing the number of generated samples generally stabilizes the estimated intervals, with the highest coverage obtained at 500 samples for most datasets. Coverage remains below the nominal 90\% level, particularly on \texttt{MATR-2} and \texttt{HUST}. For reference, using a wider 95\% interval with 500 samples increases the coverage to 90.5\%, 72.5\%, 76.1\%, 89.6\%, and 88.4\% on \texttt{MATR-1}, \texttt{MATR-2}, \texttt{HUST}, \texttt{MIX-20}, and \texttt{MIX-100}, respectively.

\subsection{RUL prediction} 
\label{sec:rul_prediction}
We evaluate three different models trained with supervised learning, diffusion processes, and flow matching, referred to as Transformer, DiffBatt, and FlowBatt, respectively. All models predict SOH trajectories from the input capacity matrix, from which the RUL is computed as the cycle index at which the SOH drops below a specified threshold (e.g., 80\% or 90\%). Thus, in contrast to the SOH task considered in \cref{sec:soh_prediction}, RUL prediction is treated here as a downstream task derived from the predicted degradation trajectories. For the probabilistic models (DiffBatt and FlowBatt), a single model is trained and evaluated by generating ten SOH trajectories from ten independent noise realizations for each input, thereby quantifying predictive variability. RUL is extracted separately from each generated trajectory before aggregating the resulting scalar predictions. For the deterministic baseline (Transformer), ten models with different random initializations are trained, and we report the mean and standard deviation of their prediction errors.
\begin{table}[!h]
\caption{\small  Comparison with baseline methods for the RUL prediction task. Best results based on RMSE are shown in bold, and second-best results are underlined. For the models introduced in this study, we report the mean and standard deviation of each error metric across ten generative draws from one trained checkpoint for FlowBatt and DiffBatt, and across ten training initializations for Transformer.}
\label{tab:rul-results}
\resizebox{\textwidth}{!}{
\small
\begin{tabular}{@{}lllllllllll@{}}
\toprule
{\multirow{2}{*}{Method}} & \multicolumn{2}{c}{\texttt{MATR-1}} & \multicolumn{2}{c}{\texttt{MATR-2}}  & \multicolumn{2}{c}{\texttt{HUST}} & \multicolumn{2}{c}{\texttt{MIX-100}} & \multicolumn{2}{c}{\texttt{MIX-20}} \\
{} &    RMSE & MAPE (\%) &     RMSE & MAPE (\%) &      RMSE & MAPE (\%) &     RMSE & MAPE (\%) &    RMSE & MAPE (\%) \\
\midrule
Training Mean                 &     399 &      28 &      511 &      36 &      420 &      18 &      573 &     59 &     593 &      102 \\
\midrule
``Variance'' Model \citep{Severson2019Data-drivenDegradation}        &     138 &      15 &      196 &      12 &        398 &      17 &      521 &      39 &     601 &      95 \\
``Discharge'' Model \citep{Severson2019Data-drivenDegradation}       &     {86} &       {8} &      \underline{173} &      \underline{11} &       {322} &   {14} &  1743 &    47 &    >2000 &      >100 \\
``Full'' Model \citep{Severson2019Data-drivenDegradation}            &     100 &      11 &      214 &      12 &         335 &      14 &      331 &      22 &     441 &      53 \\
Ridge Regression \citep{attia2021statistical}      &     125 &      13 &      188 &      11 &       1047 &      36 &      395 &     30 &     806 &      150 \\
PCR \citep{attia2021statistical}                   &     100 &      11 &      176 &      11 &      435 &      19 &      384 &      28 &     701 &      78 \\
PLSR \citep{attia2021statistical}                  &      97 &      10 &      193 &      11 &      431 &      18 &      371 &      26 &     543 &      77 \\
SVM Regression \citep{Zhang2025}       &     140 &      15 &      300 &      18 &     344 &      16 &      257 &      18 &     438 &      46 \\
Random Forest \citep{attia2021statistical}         &     140 &      15 &      202 &      11 &      348 &      16 &     {211} &   {14} &     {288} &      {31} \\
\midrule
MLP \citep{attia2021statistical}                   &   162 $\pm$ 7 &    12 $\pm$ 0 &    207 $\pm$ 4 &    11 $\pm$ 0 &   444 $\pm$ 5 &    18 $\pm$ 1 &  455 $\pm$ 37 &    27 $\pm$ 1 &   532 $\pm$ 25 &    61 $\pm$ 6\\
LSTM                  &  123 $\pm$ 11 &    12 $\pm$ 2 &   226 $\pm$ 36 &    14 $\pm$ 2 &    442 $\pm$ 32 &    20 $\pm$ 1 &   266 $\pm$ 11 &    15 $\pm$ 1 & 417 $\pm$ 62 &    37 $\pm$ 7 \\
CNN \citep{attia2021statistical}                   &  115 $\pm$ 96 &     9 $\pm$ 6 &  237 $\pm$ 107 &    17 $\pm$ 8 &    445 $\pm$ 35 &    21 $\pm$ 1  &  261 $\pm$ 38 &    15 $\pm$ 1 &  785 $\pm$ 132 &    41 $\pm$ 4 \\

BatLiNet \citep{Zhang2025} &    \textbf{59 $\pm$ 2} &     \textbf{6 $\pm$ 0} &   \textbf{163 $\pm$ 12} &    \textbf{11 $\pm$ 1} &    \underline{264 $\pm$ 9} &    \underline{10 $\pm$ 1} &   \textbf{158 $\pm$ 7} &    \textbf{10 $\pm$ 0} &   \textbf{201 $\pm$ 18} &    \textbf{18 $\pm$ 1}\\
\midrule
Transformer (ours) & 85 $\pm$ 10 & 8 $\pm$ 1 & 226 $\pm$ 39 & 14 $\pm$ 2 & 295 $\pm$ 18 & 14 $\pm$ 1 & 221 $\pm$ 34 & 15 $\pm$ 3 & 312 $\pm$ 40 & 27 $\pm$ 5 \\
DiffBatt (ours) & 68 $\pm$ 3 &6 $\pm$ 0& 202 $\pm$ 5 &16 $\pm$ 0& 282 $\pm$ 16 &10 $\pm$ 1& 218 $\pm$ 17 &13 $\pm$ 1& 295 $\pm$ 24 &30 $\pm$ 2\\
FlowBatt (ours) & \underline{67 $\pm$ 2} & \underline{6 $\pm$ 0} & 175 $\pm$ 4 & 11 $\pm$ 0 & \textbf{222 $\pm$ 12} & \textbf{9 $\pm$ 0} & \underline{179 $\pm$ 6} & \underline{11 $\pm$ 0} & \underline{229 $\pm$ 12} & \underline{18 $\pm$ 1} \\
\bottomrule
\end{tabular}
}
\end{table}

The RUL prediction results are summarized in \cref{tab:rul-results}. FlowBatt shows strong and consistent performance across all datasets, achieving the best result on \texttt{HUST} ($222\pm12$ cycles, 9\% MAPE) and second-best results on \texttt{MATR-1} ($67\pm2$), \texttt{MIX-100} ($179\pm6$), and \texttt{MIX-20} ($229\pm12$). This demonstrates that FlowBatt remains highly competitive on the downstream RUL task, despite being primarily designed for full SOH trajectory modeling rather than direct endpoint prediction.
The comparison with BatLiNet \citep{Zhang2025} is particularly relevant, as BatLiNet is specifically optimized for RUL prediction. A model designed and trained specifically for RUL may therefore perform better on this scalar task than a model trained for complete SOH trajectory prediction. On \texttt{MATR-2}, FlowBatt achieves an RMSE of $175\pm4$, close to BatLiNet ($163\pm12$), while outperforming DiffBatt ($202\pm5$) and Transformer ($226\pm39$). More generally, FlowBatt is the only model besides BatLiNet that consistently ranks among the top three across all datasets, indicating strong robustness and generalization.

Overall, these results suggest that flow-matching-based generative modeling is effective not only for SOH trajectory prediction, but also for RUL estimation derived from predicted degradation trajectories, even without being specialized exclusively for RUL.

\begin{figure}[h]
  \centering
  \includegraphics[width=\textwidth]{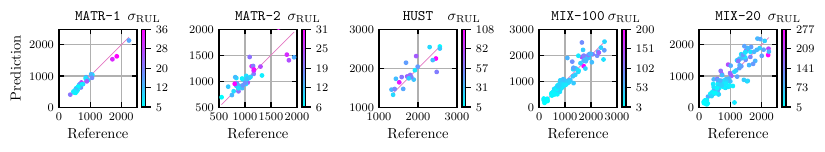}
  \caption{\small Results obtained from FlowBatt for RUL prediction. Colorbar shows the standard deviation $\sigma_{\rm RUL}$ of the predicted RUL from ten generated SOH trajectories per sample.}
  \label{fig:rul}
\end{figure}

\Cref{fig:rul} presents the RUL prediction results for all test samples across datasets using the FlowBatt model. FlowBatt provides uncertainty estimates by reporting the standard deviation of RUL values computed from ten generated trajectories per sample. The results suggest a general tendency for samples with larger prediction errors to also exhibit higher predictive variance.

\subsection{Battery degradation synthesis}

Conducting comprehensive battery degradation experiments in a real-world setting is often time-consuming and resource-intensive, requiring years to complete. Synthesizing battery degradation data offers significant advantages by reducing the time and cost of long-term experiments and ensuring consistent, high-quality data. FlowBatt can generate battery degradation curves, enabling the creation of large datasets that can enhance model training and robustness. To evaluate the effectiveness of our approach, we utilize the FlowBatt model trained on the \texttt{MATR-1} dataset to generate five synthetic SOH curves for each sample in the training set, using varying rates for classifier-free diffusion guidance. The guidance scale is denoted by $w$, with $w=1$ corresponding to standard conditional generation \citep{diffusionguidance}. For this synthesis setting, the early-cycle capacity conditioning vector is dropped with probability 0.2 during FlowBatt training; for the prediction experiments, the conditioning dropout probability is 0.0. We then leverage these synthesized SOH curves to train a random forest (RF) model for predicting RUL, while adding Gaussian noise with standard deviation equal to 10\% of the capacity-matrix standard deviation for data augmentation. To assess the quality and diversity of our generated samples, we use FID \citep{fid}, which provides insight into the similarity between real and synthetic samples, Recall \citep{recall} as the main metric for measuring diversity, and the RMSE and MAPE of the random forest model on the test dataset, serving as measures of its predictive accuracy. We also report Precision \citep{recall} for completeness. Note that FID is computed using principal component analysis (PCA) to map the data into a latent space instead of the Inception model. Results are summarized in \cref{tab:synthesis} against a random forest model trained without synthetic data following \cite{Zhang2023BatteryML:AnDegradation}. The synthetic data generated with varying guidance strengths are depicted in \cref{fig:syn}. 
\begin{table}[h]
  \centering
  \caption{\small  Results for the \texttt{MATR-1} dataset obtained using random forests trained on synthetic data generated with different diffusion guidance $w$, against a random forest trained without synthetic data following \cite{Zhang2023BatteryML:AnDegradation}. Best values per RUL error metrics, determined before rounding, are shown in bold.
  }
  \label{tab:synthesis}
  \footnotesize
  \begin{tabular}{@{}l|c|ccccc@{}}
  \toprule
  \multicolumn{2}{r}{} & \multicolumn{5}{c}{\large{$w$}} \\
  \cline{3-7}\\[-1.5ex]
  \textbf{Metric} & RF &
  $0.5$ &
  $1.0$ &
  $2.0$ &
  $3.0$ &
  $4.0$ \\
  \midrule
  FID ($\downarrow$) & NA & $0.129$ & $0.001$ & $0.001$ & $0.002$ & $0.003$ \\
  Precision ($\uparrow$) & NA & 0.97 & 1.00 & 1.00 & 1.00 & 1.00 \\
  Recall ($\uparrow$) & NA & 0.95 & 0.98 & 0.59 & 0.56 & 0.54 \\
  RUL RMSE ($\downarrow$) & 136 &
  119 &
  \textbf{101} &
  102 &
  104 &
  107 \\
  RUL MAPE (\%) ($\downarrow$) & 12 &
  11 &
  \textbf{10} &
  10 &
  10 &
  10 \\
\bottomrule
\end{tabular}
\end{table} 

On the one hand, the FID scores indicate high similarity between the real and synthetic data across the guided settings. The FID decreases from $0.129$ at $w=0.5$ to $0.001$ for $w=1.0$ and $w=2.0$. It then increases to $0.003$ at $w=4.0$. On the other hand, Recall represents the diversity by measuring the fraction of the training data manifold covered by the generated samples. The highest Recall is obtained with $w=1.0$, while increasing the guidance strength further reduces diversity, leading to a Recall of 0.54 for $w=4.0$. Precision reaches 1.00 for all settings with $w\geq1.0$. The RUL prediction metrics provide insight into the predictive performance of the random forest model. The lowest RMSE of 101 cycles and lowest MAPE of 10\% are achieved with $w=1.0$, corresponding to approximately 26\% lower RMSE and 19\% lower MAPE than the random forest model trained without synthetic data. All synthetic-data settings improve over the random forest model trained without synthetic data, which obtains an RMSE of 136 cycles and a MAPE of 12\%.

We additionally compare FlowBatt synthetic data with a naive augmentation baseline that adds Gaussian noise to capacity matrices and SOH trajectories (Gaussian jitter). We also include MLP as a second downstream predictor. The setup and results in \cref{app:synthesis_controls} show lower reported RUL RMSE with FlowBatt synthetic data for both models, with a larger reduction for RF than for MLP.

\section{Conclusions and outlook}
\label{sec:conclusions}

Battery degradation remains a major challenge for reliable and sustainable energy-storage systems. In this work, we introduced FlowBatt, a use-inspired adaptation of flow matching with a diffusion transformer backbone for probabilistic battery degradation trajectory modeling. Rather than predicting only an endpoint such as RUL, FlowBatt learns full SOH trajectories conditioned on early-cycle capacity data, providing a trajectory-level representation of battery aging and a unified model for forecasting, uncertainty estimation, and synthesis. Across all five benchmark datasets, FlowBatt obtains the best trajectory-wise SOH prediction results. Since the model generates multiple plausible degradation trajectories for the same early-life input, it also provides empirical prediction intervals that capture predictive uncertainty. In the downstream RUL task, RUL values derived from FlowBatt trajectories remain competitive with specialized RUL baselines, including BatLiNet \citep{Zhang2025}, and rank among the top methods across all evaluated datasets. The synthesis experiments further show that FlowBatt-generated SOH trajectories can improve downstream RUL modeling, reducing random-forest RMSE and MAPE on \texttt{MATR-1} by approximately 26\% and 19\%, respectively.

\textbf{Limitations.} Prediction remains challenging on \texttt{MATR-2} and \texttt{HUST}, as reflected in higher RUL errors than on \texttt{MATR-1} and more pronounced undercoverage of the empirical prediction intervals. The model conditions only on capacity matrices from fixed 20- or 100-cycle windows, without direct temperature, current-profile, charging-protocol, environmental, or cooling inputs. The fixed benchmark splits do not establish generalization to unseen chemistries or cycling protocols. FlowBatt also does not explicitly enforce electrochemical, thermal, or conservation constraints.

\textbf{Outlook.} These results support flow matching as a promising framework for probabilistic battery degradation modeling. Future work should improve uncertainty calibration, especially under distribution shifts across chemistries and cycling protocols; incorporate additional operating and environmental variables; and connect generative trajectory models more closely with physics-based constraints and descriptions of battery aging. Generalization should be assessed by holding out entire chemistries or cycling protocols. Future work should include controlled evaluation of conditioning-history length and explore encoder architectures capable of processing variable-length cycle sequences. Scaling this framework to broader battery datasets may also provide a path toward more general foundation models for battery health prediction.

\begin{ack}
This research was conducted within the Research Training Group CircularLIB, supported by the Ministry of Science and Culture of Lower Saxony with funds from the program zukunft.niedersachsen of the Volkswagen Foundation (MWK \textbar{} ZN3678). This work was also supported by the Lower Saxony Digital Science Support Space (DS\textsuperscript{3}) project as part of Hochschule.digital Niedersachsen, funded by zukunft.niedersachsen. We would like to thank the anonymous reviewers who contributed to the improvement of our manuscript with their valuable suggestions and comments.
\end{ack}

\clearpage

\setlength{\bibsep}{5pt}
\bibliographystyle{abbrvnat}
{\small \bibliography{cameraready}}

\newpage
\appendix

\section{Appendix}

\begingroup
\small
\hypersetup{linkcolor=blue}
\etocsetnexttocdepth{subsubsection}
\etocsettocstyle{\subsection*{Appendix contents}}{}
\setlength{\parskip}{1pt}
\localtableofcontents
\endgroup

\subsection{Datasets}
\label[appsec]{app:data}

\begin{figure}[h]
  \centering
  \includegraphics[width=\textwidth]{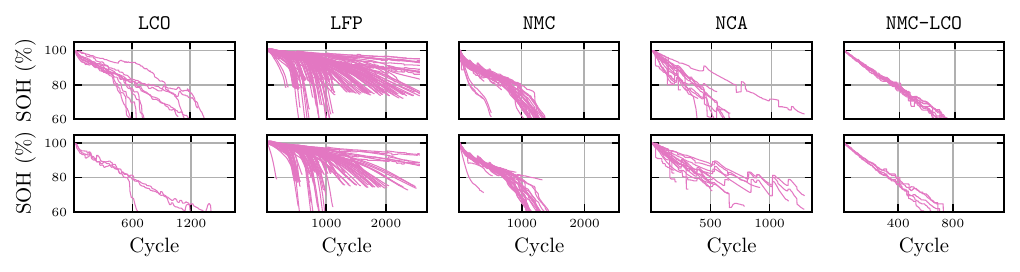}
  \caption{\small Train (up) and test (bottom) samples for each cell chemistry. The data is scaled using the SOH of the first cycle.}
  \label{fig:data}
\end{figure}

Battery degradation curves utilized for training and testing the models are depicted in \cref{fig:data} for each cell chemistry. The benchmark settings follow BatLiNet \citep{Zhang2025} with fixed train/test cell splits. The final train/test battery counts used in our experiments are summarized in \cref{tab:dataset_summary}. For \texttt{MIX-20}, our training set retains 49 cells that reach 90\% SOH before 100 cycles, which are excluded by BatLiNet. The resulting training set comprises 256 cells, compared with BatLiNet's 207, while the test set remains unchanged at 147 cells.

\begin{table}[h]
\centering
\small
\caption{\small Benchmark composition, conditioning horizon, and final train/test battery counts.}
\label{tab:dataset_summary}
\begin{tabular}{lllll}
\toprule
 Benchmark & Source data & Chemistry & Early cycles & Train/test \\
\midrule
\texttt{MATR-1} & \texttt{MATR-1} & LFP/graphite & 100 & 41/42 \\
\texttt{MATR-2} & \texttt{MATR-2} & LFP/graphite & 100 & 41/40 \\
\texttt{HUST} & \texttt{HUST} & LFP/graphite & 100 & 55/22 \\
\texttt{MIX-20} & Combined & LCO, LFP, NCA, NMC, NMC--LCO & 20 & 256/147 \\
\texttt{MIX-100} & Combined & LCO, LFP, NMC, NMC--LCO & 100 & 205/137 \\
\bottomrule
\end{tabular}%
\end{table}

The \texttt{CALCE} dataset includes full lifecycle data from 13 batteries with an LCO cathode. Each battery has a nominal capacity of 1100 mAh. They were all charged using a constant current/constant voltage protocol: 0.5C current until reaching \SI{4.2}{\volt}, maintaining \SI{4.2}{\volt} until the current dropped below \SI{0.05}{\ampere}, and a cutoff voltage of \SI{2.7}{\volt} \citep{CALCE1, CALCE2}. The C-rate expresses the charging or discharging current relative to nominal capacity, with 1C corresponding to a current that would transfer the nominal capacity in one hour.

The \texttt{MATR} dataset, provided by \citet{Severson2019Data-drivenDegradation} and \citet{CLO}, is one of the largest public datasets containing 180 commercial 18650 LFP batteries. These batteries, cycled at a forced convection temperature chamber of \SI{30}{\degreeCelsius}, have a nominal capacity of 1.1 Ah and a nominal voltage of \SI{3.3}{\volt}. The dataset comprises three subsets: \texttt{MATR-1}, \texttt{MATR-2} \citep{Severson2019Data-drivenDegradation}, and \texttt{CLO} \citep{CLO}, all categorized due to distinct measurement batches. 

The \texttt{HUST} dataset includes 77 LFP batteries, similar to those in the \texttt{MATR} dataset. These batteries followed an identical charging protocol with varying multi-stage discharge protocols, all conducted at a constant temperature of \SI{30}{\degreeCelsius} \citep{HUST}. 

The \texttt{HNEI} dataset contains 14 commercial 18650 cells with a graphite anode and a blended NMC and LCO cathode. These cells were cycled at 1.5C to 100\% depth of discharge for over 1000 cycles at room temperature \citep{HNEI}. 

The \texttt{SNL} dataset includes 61 commercial 18650 cells (NCA, NMC, and LFP), cycled to 80\% capacity. The study evaluates the impact of temperature, depth of discharge, and discharge current on long-term degradation \citep{SNL}. 

The \texttt{UL\_PUR} dataset comprises 10 commercial pouch cells with a graphite negative electrode and an NCA cathode. These cells were cycled at 1C between \SI{2.7}{\volt} and \SI{4.2}{\volt}, equivalent to 0-100\% state of charge (SOC), at room temperature until reaching 10-20\% capacity fade. Additionally, modules were cycled at C/2 between \SI{13.7}{\volt} and \SI{21.0}{\volt} until 20\% capacity fade \citep{UL_PUR, UL_PUR2}. 

The \texttt{RWTH} dataset contains data from 48 lithium-ion battery cells aged under identical conditions. These cells feature a carbon anode and an NMC cathode \citep{RWTH}. The cells were cycled at a constant ambient temperature of \SI{25}{\degreeCelsius}. Each cycle involved a 30-minute discharge phase down to \SI{3.5}{\volt} and a 30-minute charge phase up to \SI{3.9}{\volt}, with the currents capped at a maximum of \SI{4}{\ampere}. This resulted in cycles between approximately 20\% and 80\% state of charge.

\textbf{Licenses and terms of use.} All benchmark datasets utilized in this study are publicly available for academic research and are curated within the open-source BatteryML framework \citep{Zhang2023BatteryML:AnDegradation}, which is released under the MIT License. The PyTorch and diffusers \citep{von-platen-etal-2022-diffusers} libraries used for our implementation are also open-source and their respective licenses (e.g., BSD-style, Apache 2.0) are strictly respected.

\subsection{Details of FlowBatt model}
\subsubsection{\texorpdfstring{Capacity matrix}{Capacity matrix}}
\label[appsec]{app:capacity_matrix}

For each battery, the discharge-capacity curve at observed cycle $i$ is interpolated onto a common grid of 100 voltage points. The capacity matrix is a cycle-by-voltage array of baseline-subtracted discharge-capacity curves, $\mQ=[M_{i,j}]$, with entries
\begin{equation}
    M_{i,j}=Q_d^{(i)}(V_j)-Q_d^{(10)}(V_j),
\end{equation}
where $i$ is the observed-cycle index, $j$ is the index on the common voltage grid, $V_j$ is the voltage at grid point $j$, and $Q_d^{(i)}(V_j)$ is the discharge capacity at voltage $V_j$ during cycle $i$. Cycle 10 is the reference curve subtracted from every observed cycle. The matrix therefore contains only baseline-subtracted discharge-capacity curves and records how the capacity--voltage response evolves during early cycling.
We selected this representation following \citet{Severson2019Data-drivenDegradation}, \citet{attia2021statistical}, and \citet{Zhang2023BatteryML:AnDegradation}, which identify early-cycle differential discharge-capacity data as informative for battery degradation and lifetime. Its voltage-dependent structure reflects changes in energy dissipation and may capture open-circuit and kinetic effects, heterogeneous aging, and early voltage signatures of degradation. In particular, \citet{attia2021statistical} reported that a model using only $\log_{10}\mathrm{Var}[\Delta Q_{100-10}(V)]$ achieved lower RMSE than their full model containing temperature and internal-resistance features on the secondary MATR test set. In the present study, FlowBatt does not receive temperature, internal resistance, current profiles, charging protocols, or environmental conditions as direct inputs. However, the conditioning pathway can be extended to incorporate these variables when they are available.

\subsubsection{Hyperparameters}
\label[appsec]{app:hyperparams}

\Cref{tab:hyperparams} summarizes the hyperparameters used for training our FlowBatt model. For implementing flow matching, we build upon the official \texttt{flow\_matching} library from Facebook Research~\citep{lipman2024flowmatchingguidecode}. For diffusion-related components, including optimizers and schedulers, we rely on the Hugging Face \texttt{diffusers} package~\citep{von-platen-etal-2022-diffusers}. 

\begin{table}[h]
\centering
\caption{\small Training and model hyperparameters.}
\label{tab:hyperparams}
\footnotesize
\begin{tabular}{ll}
\toprule
Category & Major hyperparameters \\
\midrule
\multirow{6}{*}{Training} 
& Epochs: 10000 \\
& Initial learning rate: $1\times 10^{-3}$ \\
& Optimizer: AdamW \\
& Learning rate scheduler: Cosine schedule with 100 warmup steps \\
\midrule
\multirow{7}{*}{Model (DiT)} 
& Sequence length: 256 \\
& Capacity matrix shape: $(1,100,100)$, or $(1,20,100)$ for \texttt{MIX-20} \\
& Number of DiT blocks: 4 \\
& Channels (embedding dim): 16 \\
& Attention heads: 1 \\
& MLP ratio: 4.0 \\
& Conditioning dropout probability: 0.0 (default), 0.2 (SOH synthesis, \cref{tab:synthesis}) \\
\midrule
\multirow{1}{*}{Flow matching} 
& Scheduler: Conditional optimal transport schedule \\
\bottomrule
\end{tabular}
\end{table}

For model development, we preprocess the SOH trajectories; we limit each trajectory to a maximum of 2560 charge-discharge cycles and subsample the SOH values every ten cycles. 
Furthermore, we truncate the trajectory once the SOH drops below 70\%, since this threshold is typically considered below the EOL criterion for many applications. 
If a trajectory reaches this SOH cutoff before 2560 cycles, we pad the remaining part of the sequence with a constant SOH value of 70\%. 
As a result, all SOH trajectories are represented as fixed-length sequences of 256 steps, which defines the input dimension of our model.

\subsubsection{Compute resources}
\label[appsec]{app:compute_resources}
All experiments, including model training, baseline evaluations, and trajectory generation, were conducted using PyTorch on a workstation equipped with a single NVIDIA RTX 3090 GPU (24 GB VRAM) and a 12th Gen Intel Core i9-12900K CPU. Training the FlowBatt model for 10,000 epochs takes approximately 30 minutes per dataset. On a single GPU, predicting one complete SOH trajectory per battery in a batch of 42 \texttt{MATR-1} test batteries takes approximately 3~ms for Transformer, 0.5~s for FlowBatt, and 2.6~s for DiffBatt, averaged over three runs following one warm-up. FlowBatt therefore achieves approximately a fivefold inference speedup over DiffBatt.

\subsubsection{DiT block with adaLN-Zero conditioning}
\label[appsec]{app:ditblock}

In both FlowBatt and DiffBatt models, each DiT block updates the tokenized SOH representation using self-attention and an MLP, while conditioning both sublayers on the flow or diffusion timestep and the early-cycle capacity matrix. The timestep embedding and CNN-based capacity embedding are combined into a global conditioning vector, which is mapped to adaptive layer-normalization shifts, scale offsets, and residual gates following adaLN-Zero \citep{PeeblesDIT}. The residual gates are initialized near zero, so the block initially behaves close to an identity map and the conditioning effect is learned gradually during training. The computational steps for one block are summarized in algorithm~\ref{alg:ditblock}.

\begin{algorithm}[h]
  \footnotesize
\caption{\small One DiT block with adaLN-Zero conditioning}
\label{alg:ditblock}
\KwIn{Hidden states $\vh \in \mathbb{R}^{T \times d}$, timestep $t$, auxiliary input $\mQ$ (capacity matrix).}
\KwOut{Updated hidden states $\vh' \in \mathbb{R}^{T \times d}$.}

\tcc{Step 1: Form conditioning vector}
$\vt_{\mathrm{emb}} \gets \phi(t)$ \hfill \tcp{timestep embedding}
$\vc_q \gets \mathrm{CNN}(\mQ)$ \hfill \tcp{auxiliary embedding}
$\vc \gets \vt_{\mathrm{emb}} + \vc_q$ \hfill \tcp{global conditioning vector}

\tcc{Step 2: Generate modulation parameters}
$(\gamma_{\mathrm{attn}}, \beta_{\mathrm{attn}}, \alpha_{\mathrm{attn}}, 
 \gamma_{\mathrm{mlp}}, \beta_{\mathrm{mlp}}, \alpha_{\mathrm{mlp}}) 
 \gets \mathrm{MLP}(\vc)$

\tcc{Step 3: Attention sublayer}
$\vz \gets \mathrm{LN}(\vh)$ \hfill \tcp{pre-norm}
$\vz \gets (1+\gamma_{\mathrm{attn}}) \odot \vz + \beta_{\mathrm{attn}}$ \hfill \tcp{adaLN modulation}
$\vh \gets \vh + \alpha_{\mathrm{attn}} \odot \mathrm{MSA}(\vz)$ \hfill \tcp{residual update}

\tcc{Step 4: MLP sublayer}
$\vz \gets \mathrm{LN}(\vh)$ \hfill \tcp{pre-norm}
$\vz \gets (1+\gamma_{\mathrm{mlp}}) \odot \vz + \beta_{\mathrm{mlp}}$ \hfill \tcp{adaLN modulation}
$\vh' \gets \vh + \alpha_{\mathrm{mlp}} \odot \mathrm{MLP}(\vz)$ \hfill \tcp{residual update}

\Return $\vh'$
\end{algorithm}

\subsection{Explaining CNN performance via Grad-CAM}
\label[appsec]{app:cnn}

\begin{figure}[h]
  \centering
  \begin{subfigure}[t]{0.49\textwidth}
   \centering
    \includegraphics[width=0.8\textwidth]{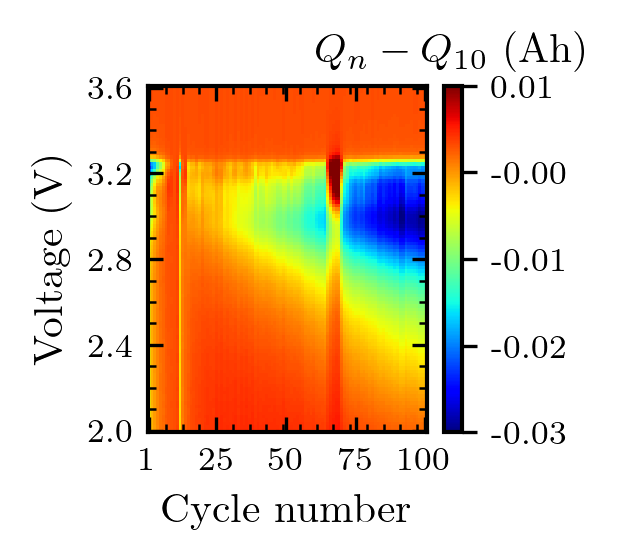}
    \caption{\small Capacity matrix}
    \label{fig:qmatrix_a}
  \end{subfigure}
  \hfill
  \begin{subfigure}[t]{0.49\textwidth}
   \centering
      \includegraphics[width=0.8\textwidth]{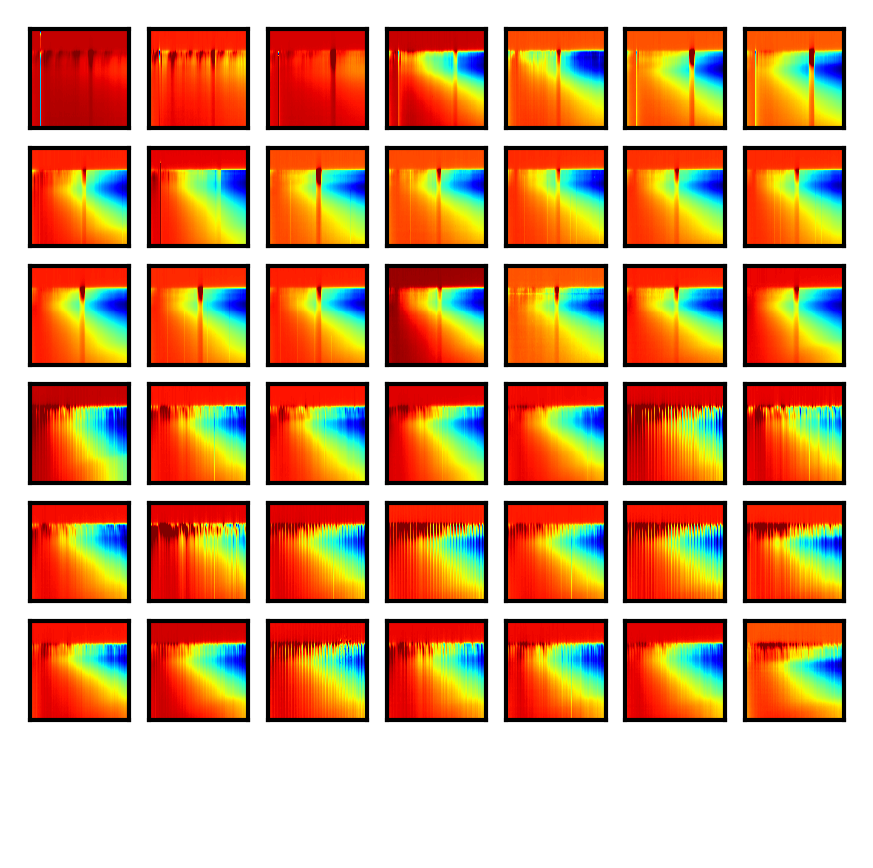}
      \caption{\small \texttt{MATR-1} test samples}
      \label{fig:qmatrix_b}
  \end{subfigure}
  \caption{\small (a) One sample of capacity matrix from \texttt{MATR-1} train dataset. (b) All test samples of \texttt{MATR-1} dataset.}
  \label{fig:qmatrix}
\end{figure}

\begin{figure}[h]
  \centering
  \begin{subfigure}[t]{0.49\textwidth}
   \centering
    \includegraphics[width=0.8\textwidth]{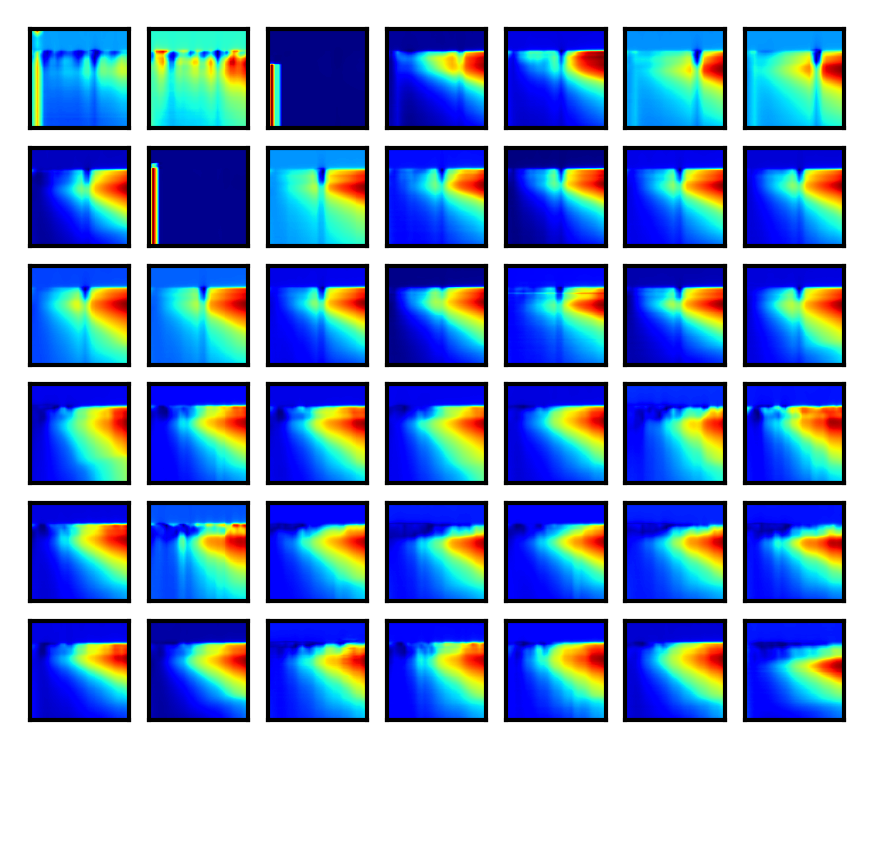}
    \caption{\small Grad-CAM heatmaps for best model}
  \end{subfigure}
  \hfill
  \begin{subfigure}[t]{0.49\textwidth}
   \centering
      \includegraphics[width=0.8\textwidth]{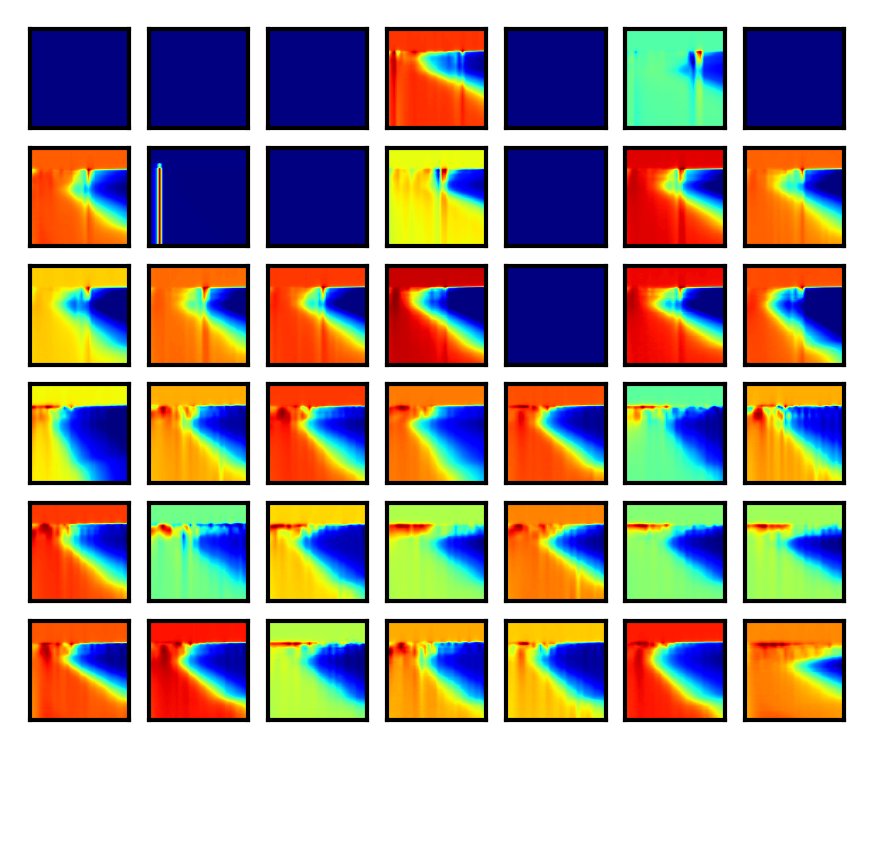}
      \caption{\small Grad-CAM heatmaps for worst model}
  \end{subfigure}
  \caption{\small Grad-CAM heatmaps for CNN model proposed in \cite{attia2021statistical,Zhang2023BatteryML:AnDegradation}.}
  \label{fig:cnngradcam}
\end{figure}

For the RUL prediction task, we implement a CNN model based on the architecture proposed in \cite{attia2021statistical,Zhang2023BatteryML:AnDegradation,Zhang2025}. The model consists of convolutional layers with ReLU activation functions, followed by a fully connected layer. The model is trained using the Adam optimizer with a learning rate of 0.001 and a batch size of 128 (this is equal to full-batch training for \texttt{MATR-1} dataset) for 1000 epochs. The model is trained via mean-squared error (MSE) as the loss function. We train ten models with different initialization seeds and report the RMSE for each model on the test dataset. The results are summarized in \cref{tab:cnn-rul} for different seeds. As observed in the table, the CNN model exhibits high variability in performance across different seeds, with RMSE values ranging from 60 to 306 cycles. This variability indicates that the model's performance is sensitive to initialization and may not consistently capture the underlying degradation patterns. This lack of robustness has been also reported in \cite{attia2021statistical,Zhang2023BatteryML:AnDegradation,Zhang2025}.
\begin{table}[h]
    \centering
    \footnotesize
    \begin{threeparttable} 
    \caption{\small RUL prediction results obtained from CNN models with different initialization seeds for \texttt{MATR-1} dataset.}
    \label{tab:cnn-rul}
    \begin{tabular}{@{}lcccccccccc@{}}
    \toprule
    \textbf{Seed} & \textbf{0} & \textbf{1} & \textbf{2} & \textbf{3} & \textbf{4} & \textbf{5} & \textbf{6} & \textbf{7} & \textbf{8} & \textbf{9} \\
    \midrule
    Benchmark CNN \citep{Zhang2023BatteryML:AnDegradation} & 76 & 67 & 64 & 74 & 60 & 82 & 65 & 79 & {367} & 78 \\
    CNN (Replicated)     & 71 & 77 & 83 & {306} & 79 & 60 & 78 & 62 & 88 & 64 \\
    XAI-CNN (Ours)  & 57 & 54 & 54 & 72 & 66 & 64 & 65 & 60 & 59 & 61\\
    \bottomrule
    \end{tabular}
    \end{threeparttable}
\end{table}
To identify the reason behind the high variability in performance, we employ Grad-CAM \citep{Selvaraju2017Grad-CAM:VisualExplanations} to visualize the regions of the input capacity matrix that the CNN model focuses on when making predictions. \Cref{fig:qmatrix_a} shows one sample of the capacity matrix from the \texttt{MATR-1} train dataset, while \cref{fig:qmatrix_b} displays all test samples of the \texttt{MATR-1} dataset. The heatmaps in \cref{fig:cnngradcam} illustrate the Grad-CAM results for the best and worst-performing CNN models based on RMSE. The heatmaps obtained from the worst model indicate that the gradients for some test samples are zero, suggesting that the model fails to perform meaningful predictions for these samples. Same observations for the training samples indicate that the model fails to learn from those training samples. Moreover, the model tends to focus on early cycles, which may not provide sufficient information for accurate RUL prediction. In contrast, the heatmaps for the best-performing model show that the model focuses on a broader range of cycles, including later ones that are more indicative of degradation trends. 

To enhance the robustness and accuracy of the CNN model, we provide a simple remedy by replacing the ReLU activation functions with Leaky ReLU activation functions with negative slope equal to 0.3. This choice is due to the realization that the ReLU activation function can lead to dead neurons, especially when the input data contains negative values or when the weights are initialized poorly. Dead neurons do not contribute to learning, which can significantly impact the model's performance and robustness. Leaky ReLU allows a small, non-zero gradient when the unit is not active, which helps to keep the neurons alive and ensures that they continue to learn during training. This modification helps to mitigate the issue of dead neurons and improves the overall robustness of the CNN model. 
\begin{figure}[h]
  \centering
  \begin{subfigure}[t]{0.49\textwidth}
   \centering
    \includegraphics[width=0.8\textwidth]{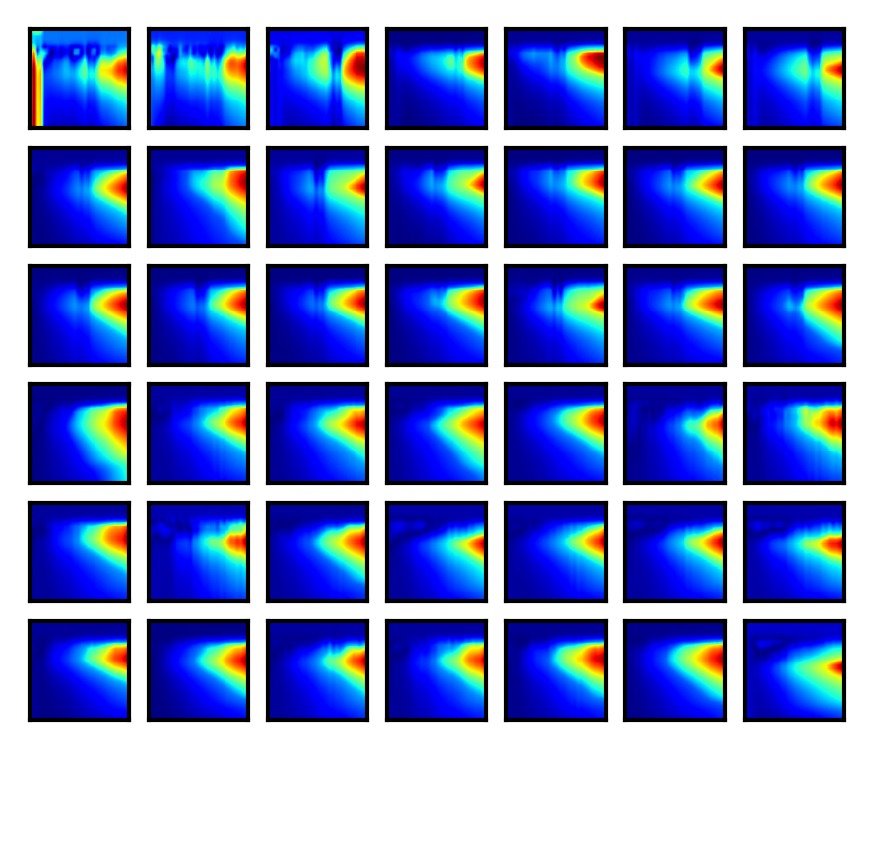}
    \caption{\small Grad-CAM heatmaps for best model}
  \end{subfigure}
  \hfill
  \begin{subfigure}[t]{0.49\textwidth}
   \centering
      \includegraphics[width=0.8\textwidth]{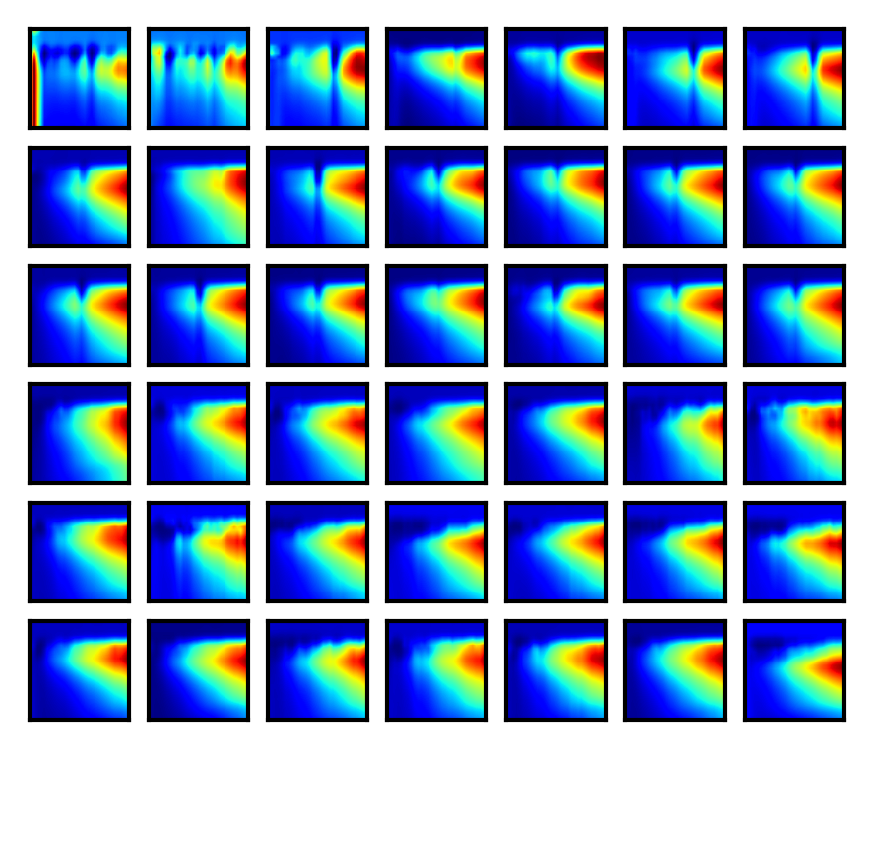}
      \caption{\small Grad-CAM heatmaps for worst model}
  \end{subfigure}
  \caption{\small Grad-CAM heatmaps for XAI-CNN model.}
  \label{fig:xcnngradcam}
\end{figure}
We refer to this modified model as XAI-CNN. The XAI-CNN model is trained using the same hyperparameters as the CNN model. The RMSE results for different seeds are summarized in \cref{tab:cnn-rul}. This model demonstrates significantly improved robustness, with RMSE values ranging from 54 to 72 cycles across different seeds, and overall provides very accurate RUL predictions with RMSE of 61$\pm$5 for \texttt{MATR-1} dataset. The Grad-CAM heatmaps for the best and worst-performing XAI-CNN models are shown in \cref{fig:xcnngradcam}. The heatmaps indicate that the XAI-CNN model consistently focuses on relevant regions of the capacity matrix across all test samples, suggesting that it effectively captures important degradation patterns. This improvement in focus likely contributes to the enhanced performance and robustness of the XAI-CNN model compared to the original CNN architecture.

We use the XAI-CNN model as the CNN encoder in FlowBatt and DiffBatt. The final linear layer of the XAI-CNN model is removed, and the output from the last convolutional layer is flattened and passed to a linear layer to match the dimension of the condition vector in FlowBatt and DiffBatt. 

\subsection{Experimental setup}
\subsubsection{BatteryML}
\label[appsec]{app:soh_BML}

\textbf{SOH prediction.}
BatteryML \citep{Zhang2023BatteryML:AnDegradation} is introduced as a general benchmark framework for battery degradation prediction, including SOH-related tasks. However, at the time of our study, the public SOH workflow was not as explicit and directly reproducible as the released RUL setting. Therefore, for the BatteryML SOH experiments, we adapted the public BatteryML framework to the SOH task while keeping the setup as close as possible to the released RUL-oriented configurations. In particular, we preserved the public dataset definitions, train-test splits, feature extractors, and transformation pipeline whenever possible, and only modified the label definition and output setting to support SOH prediction under our cycle-wise and trajectory-wise evaluation protocols.

The RF and MLP baselines of \citet{Zhang2023BatteryML:AnDegradation} use the baseline-subtracted discharge-capacity curve from the final early-life cycle, rather than the complete capacity matrix defined in \cref{app:capacity_matrix}.

\begin{table}[h]
\centering
\caption{\small Training and model settings for the BatteryML baselines used in the SOH experiments.}
\label{tab:batteryml-soh-config}
\footnotesize
\begin{tabular}{ll}
\toprule
Category & Major hyperparameters \\
\midrule
Evaluation & Cycle-wise SOH \\
& Trajectory-wise SOH\\
\midrule
MLP & Input representation: capacity matrix \\
& Input settings: final early-life curve minus baseline curve \\
& Target: cycle-wise SOH or SOH trajectory \\
& Trajectory settings: max cycles $=2560$, stride $=10$, SOH cutoff $=0.7$ \\
& Label transformation: Z-score \\
& Optimizer: Adam, learning rate $=10^{-3}$ \\
& Epochs: 5000, batch size: 128 \\
& Activation: \texttt{tanh}, dropout: 0.05 \\
\midrule
Random Forest & Input representation: capacity matrix \\
& Input settings: final early-life curve minus baseline curve \\
& Target: cycle-wise SOH or SOH trajectory \\
& Trajectory settings: max cycles $=2560$, stride $=10$, SOH cutoff $=0.7$ \\
& Label transformation: Z-score \\
& Estimator: default Random Forest training settings \\
\bottomrule
\end{tabular}
\end{table}

We evaluated the RF and MLP baselines of \citet{Zhang2023BatteryML:AnDegradation} under the same dataset definitions and train-test splits used throughout this study. For cycle-wise SOH prediction, evaluation was performed at cycle indices 100, 200, and 300. For trajectory-wise SOH prediction, the model output was evaluated over the full SOH trajectory. We additionally evaluated full-capacity-matrix variants for trajectory prediction. RF uses the flattened matrix, whereas MLP processes the voltage and cycle dimensions sequentially. For these variants, hyperparameters were selected using only training data, through three-fold cross-validation for RF and an internal 80/20 split for MLP, followed by refitting on the complete training split. For all RF and MLP models, we report the mean and standard deviation over ten random seeds.
FlowBatt achieves lower trajectory-wise RMSE than both full-matrix baselines on all five datasets (\cref{tab:soh-trajectorywise,tab:full-matrix-baselines}).

\begin{table}[h]
\centering
\caption{\small Trajectory-wise SOH RMSE (\%) for RF and MLP using the final early-life curve or the complete capacity matrix. Values are mean $\pm$ standard deviation over ten training runs. Final-cycle results are reproduced from \cref{tab:soh-trajectorywise}.}
\label{tab:full-matrix-baselines}
\footnotesize
\setlength{\tabcolsep}{4pt}
\begin{tabular}{@{}lccccc@{}}
\toprule
Method & \texttt{MATR-1} & \texttt{MATR-2} & \texttt{HUST} & \texttt{MIX-20} & \texttt{MIX-100} \\
\midrule
RF (final-cycle curve) & 2.01 $\pm$ 0.04 & 2.50 $\pm$ 0.01 & 3.57 $\pm$ 0.03 & 3.42 $\pm$ 0.02 & 5.45 $\pm$ 0.04 \\
MLP (final-cycle curve) & 1.97 $\pm$ 0.11 & 3.06 $\pm$ 0.17 & 4.11 $\pm$ 0.13 & 2.23 $\pm$ 0.09 & 3.13 $\pm$ 0.14 \\
\midrule
RF (full matrix) & 2.26 $\pm$ 0.02 & 3.05 $\pm$ 0.06 & 4.20 $\pm$ 0.03 & 1.50 $\pm$ 0.01 & 2.51 $\pm$ 0.02 \\
MLP (full matrix) & 2.75 $\pm$ 0.08 & 4.04 $\pm$ 0.35 & 4.16 $\pm$ 0.23 & 2.87 $\pm$ 0.20 & 2.91 $\pm$ 0.14 \\
\bottomrule
\end{tabular}
\end{table}

\textbf{RUL prediction.}
For RUL prediction, we follow the implementation and benchmark settings of \citet{Zhang2023BatteryML:AnDegradation}.

\subsubsection{BatLiNet}
\label[appsec]{app:soh_BatLinet}

\textbf{SOH prediction.}
BatLiNet \citep{Zhang2025} was originally designed for battery lifetime prediction, i.e., the RUL task, across diverse aging conditions. To obtain SOH prediction results in our study, we adapted the original BatLiNet framework with minimal task-specific modifications. In particular, we replaced the original RUL target with an SOH target while preserving the same benchmark datasets, train-test splits, feature extraction pipeline, and model hyperparameters as closely as possible. This allowed us to evaluate BatLiNet under both cycle-wise and trajectory-wise SOH settings while keeping the experimental protocol aligned with the original benchmark design.

\begin{center}
\captionof{table}{Training and model settings for the BatLiNet baseline used in the SOH experiments.}
\label{tab:batlinet-hyperparams}
\footnotesize
\setlength{\tabcolsep}{4pt}
\begin{tabular}{@{}p{0.22\textwidth}p{0.70\textwidth}@{}}
\toprule
Category & Major hyperparameters \\
\midrule
Evaluation 
& Cycle-wise SOH; trajectory-wise SOH \\
\midrule
Model 
& Input channels: 6; hidden channels: 32; input size: $100 \times 1000$; feature extractor: BatLiNetFeatureExtractor, \texttt{smooth\_features=False} \\
\midrule
Training 
& Epochs: 1000; train batch size: 128; test batch size: 1; train support size: 2; test support size: 32; gradient accumulation steps: 1 \\
\midrule
Target setup 
& Original BatLiNet uses RUL labels with log-scale and Z-score transformation. For SOH experiments, only the target label was adapted while keeping the remaining setup unchanged. \\
\bottomrule
\end{tabular}
\end{center}

For cycle-wise SOH prediction, evaluation was performed at cycle indices 100, 200, and 300. For trajectory-wise SOH prediction, the full predicted SOH trajectory was evaluated. In all BatLiNet \citep{Zhang2025} SOH experiments, we report the mean and standard deviation over 10 random seeds.

\textbf{RUL prediction.}
For the RUL task, we follow the original BatLiNet setting \citep{Zhang2025}.

\subsection{Additional SOH results}
\label[appsec]{app:soh_results_additional}

\subsubsection{Synthetic SOH trajectory samples}
\label[appsec]{app:synthesis_controls}

To assess whether naive augmentation can provide similar gains, we compare Gaussian jitter with FlowBatt synthetic data for downstream RUL prediction on the \texttt{MATR-1} dataset. In addition to RF, we evaluate an MLP that takes the complete capacity matrix as input and processes its voltage and cycle dimensions sequentially. Gaussian jitter applies independent pointwise Gaussian noise to the capacity matrix and SOH trajectory, with standard deviation equal to 10\% of the corresponding training-data standard deviation. RUL is then extracted from the perturbed trajectory at the first 80\% SOH crossing.

\begin{table}[htbp]
  \centering
  \caption{\small RUL RMSE (cycles) for real-only training and data augmentation on \texttt{MATR-1}. Published RF and MLP benchmark results from \citet{Zhang2025} are included for reference. Values include the reported standard deviations where available. The lowest reported RMSE for each downstream model is shown in bold.}
  \label{tab:synthesis-controls}
  \footnotesize
  \setlength{\tabcolsep}{6pt}
  \begin{tabular}{@{}lcccc@{}}
  \toprule
  Model & Published benchmark & Real only & Gaussian jitter & \shortstack{FlowBatt\\synthetic data} \\
  \midrule
  RF & 140 & 136 $\pm$ 3 & 110 $\pm$ 1 & \textbf{101 $\pm$ 2} \\
  MLP & 162 $\pm$ 7 & 88 $\pm$ 7 & 85 $\pm$ 4 & \textbf{79 $\pm$ 4} \\
  \bottomrule
  \end{tabular}
\end{table}

As shown in \cref{tab:synthesis-controls}, Gaussian jitter improves the reported RUL RMSE for both downstream models, while FlowBatt synthetic data yields the lowest errors. The additional reduction is larger for RF and more modest for MLP. These results support the utility of synthetic data for the evaluated downstream models and augmentation settings. The RF RMSE of 140 cycles in \cref{tab:rul-results} is the benchmark result reported by \citet{Zhang2025}, whereas the 136-cycle baseline in \cref{tab:synthesis,tab:synthesis-controls} is our independent reproduction. This difference reflects separate implementations and runs, not a gain from augmentation.

\begin{figure}[th]
  \centering
  \includegraphics[width=\textwidth]{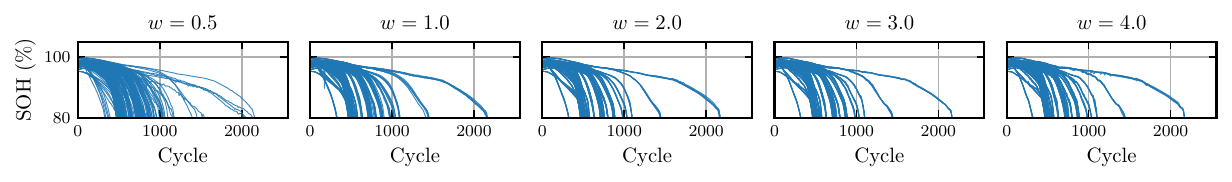}
  \caption{\small \texttt{MATR-1} synthetic data generated by FlowBatt with different guidance strengths.}
  \label{fig:syn}
\end{figure}

\Cref{fig:syn} shows synthetic SOH trajectories generated by FlowBatt on the \texttt{MATR-1} dataset for different classifier-free guidance strengths. Across all settings, the generated curves preserve the expected monotonic degradation behavior while covering a broad range of lifetimes and fading rates. Increasing the guidance strength changes the spread of the generated trajectories, with higher values producing a more concentrated set of samples, consistent with the reduction in Recall reported in \cref{tab:synthesis}.

\subsubsection{Trajectory-wise SOH prediction}
\label[appsec]{app:soh_fbatt}

For prediction tasks we generate ten samples for each input capacity matrix. The capacity matrix is constructed from the first 100 or 20 cycles. The RMSE for an SOH sample $j$ is computed as
\begin{equation}
  {\rm RMSE}_j = \sqrt{\frac{1}{n_j}\sum_{i=1}^{n_j}({\tilde{y}}_i - {y}_i)^2}
\end{equation} 
where $\tilde{y}$ and $y$ represent the predicted and the reference SOH in percentage, respectively, $i$ denotes the cycle number and $n_j$ is the cycle number at which the predicted SOH reaches the EOL. Further, we report the mean RMSE across all the test samples as the RMSE for the dataset. We pad both the reference and predicted sequences with the last value to ensure they have the same length. The mean and standard deviation of the RMSE are computed over ten generated samples per each input capacity matrix.

For downstream RUL evaluation, the scalar target is the first cycle at which SOH reaches or falls below 80\% (90\% for \texttt{MIX-20}). We report RMSE in cycles and MAPE (\%).

\begin{figure}[h]
  \centering
  \includegraphics[width=0.8\textwidth]{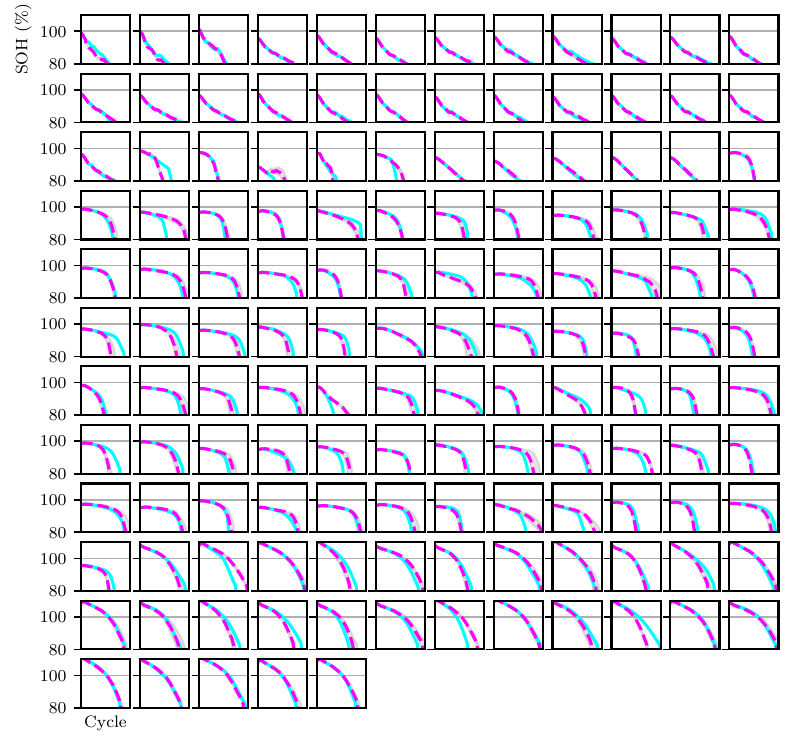}
  \caption{\small SOH predictions obtained from FlowBatt against reference for all the test samples of \texttt{MIX-100} dataset. The pink dashed line shows the prediction and the cyan solid line shows the reference. The gray area shows the prediction uncertainty computed from ten generated samples.}
  \label{fig:mixflowbatt}
\end{figure}

\Cref{fig:mixflowbatt} illustrates the predicted SOH versus the reference SOH for all test samples in the \texttt{MIX-100} dataset. The results demonstrate that FlowBatt effectively captures various degradation dynamics and accurately predicts SOH for the majority of test samples and highlights FlowBatt's ability to generalize across different battery chemistries and operational conditions present in the \texttt{MIX-100} dataset. This capability is essential for developing reliable battery health monitoring systems that can adapt to diverse usage patterns and environmental factors. 

\begin{figure}[h]
  \centering
  \includegraphics[width=0.8\textwidth]{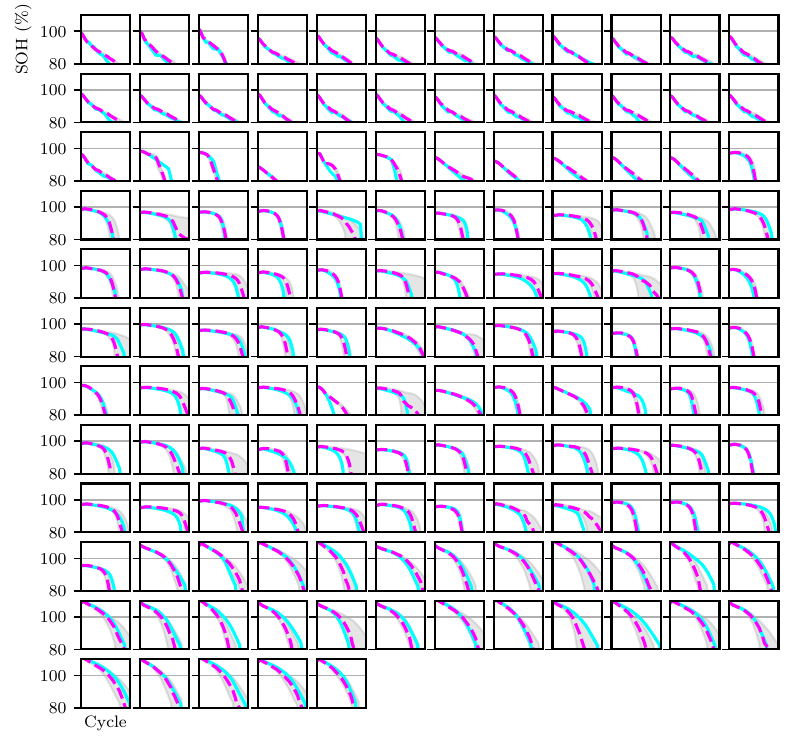}
  \caption{\small SOH predictions obtained from DiffBatt against reference for all the test samples of \texttt{MIX-100} dataset. The pink dashed line shows the prediction and the cyan solid line shows the reference. The gray area shows the prediction uncertainty computed from ten generated samples.}
  \label{fig:mixdiffbatt}
\end{figure}

\begin{figure}[h]
  \centering
  \includegraphics[width=0.8\textwidth]{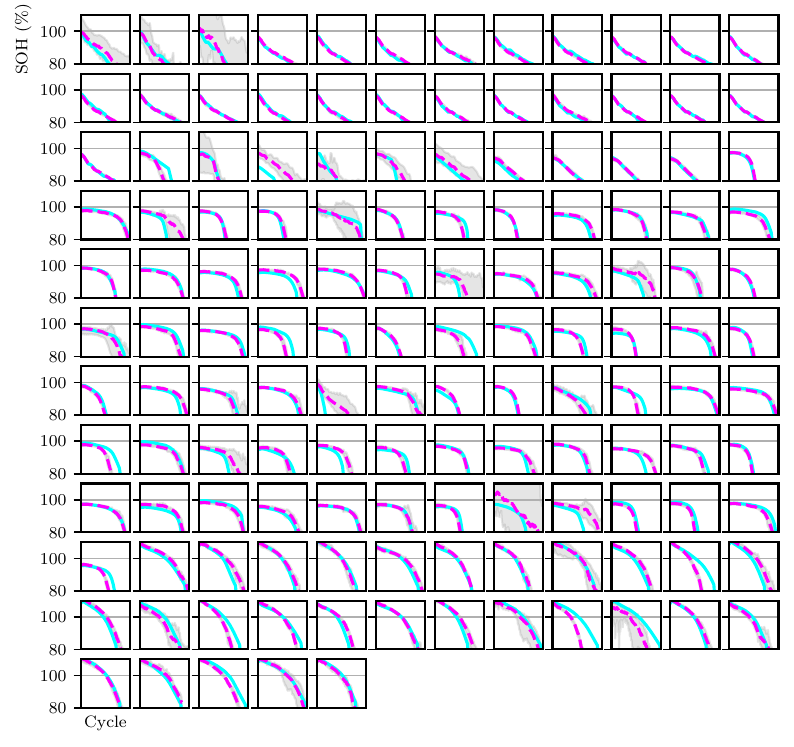}
  \caption{\small SOH predictions obtained from Transformer model against reference for all the test samples of \texttt{MIX-100} dataset. The pink dashed line shows the prediction and the cyan solid line shows the reference. The gray area shows the variability across predictions from ten independently trained Transformer models.}
  \label{fig:mixtrf}
\end{figure}

\Cref{fig:mixdiffbatt} and \cref{fig:mixtrf} show the predicted SOH against the reference SOH for all test samples in the \texttt{MIX-100} dataset obtained from DiffBatt and Transformer, respectively. Compared to FlowBatt, DiffBatt shows a higher deviation from the reference for several test samples. The Transformer model shows a higher deviation from the reference for most of the test samples compared to both FlowBatt and DiffBatt. These results further highlight the advantages of flow matching over both diffusion-based training and supervised learning.

\subsubsection{Cycle-wise SOH prediction}
\label[appsec]{app:soh_cyclewise}

For completeness and comparability with prior BatteryML-style evaluation \citep{Zhang2023BatteryML:AnDegradation}, we also consider a \emph{cycle-wise} SOH prediction setting. In this setting, the model predicts the SOH at a predefined cycle index, and the error is reported as the RMSE across all cells in the dataset. Since BatteryML does not natively evaluate full degradation trajectories for SOH, we adapt both FlowBatt and BatLiNet \citep{Zhang2025} to the same cycle-wise setting in order to ensure a fair comparison with the BatteryML baselines.

In the main comparison, cycle-wise SOH is evaluated at cycle 100, which follows the standard BatteryML-style early-life setting \citep{Zhang2023BatteryML:AnDegradation}. To further assess whether the relative behavior of the models remains consistent at later stages of degradation, we additionally extend the cycle-wise analysis to cycle indices 200 and 300. This extended evaluation provides a complementary pointwise view of SOH prediction beyond the early-cycle regime.

\begin{table}[h]
  \centering
  \caption{\small Cycle-wise SOH prediction results at cycle 100, reported as RMSE (\%) across cells. Best results are shown in bold and second-best results are underlined. This table follows the BatteryML-style cycle-wise SOH protocol, where SOH is predicted at a predefined cycle index and RMSE is computed across cells. These results are therefore not directly comparable to the trajectory-wise SOH prediction results in \cref{tab:soh-trajectorywise}.}
  \label{tab:soh-cyclewise}
  \footnotesize
  \begin{tabular}{@{}lccccc@{}}
  \toprule
  Method & \texttt{MATR-1} & \texttt{MATR-2} & \texttt{HUST} & \texttt{MIX-20} & \texttt{MIX-100} \\
  \midrule
  BatLiNet & \underline{0.33 $\pm$ 0.01} & \textbf{0.18 $\pm$ 0.01} & \textbf{0.05 $\pm$ 0.03} & \textbf{0.33 $\pm$ 0.01} & \underline{0.27 $\pm$ 0.02} \\
  RF (BatteryML) & 0.88 $\pm$ 0.01 & 1.52 $\pm$ 0.03 & 1.18 $\pm$ 0.03 & 2.19 $\pm$ 0.05 & 2.04 $\pm$ 0.07 \\
  MLP (BatteryML) & 1.01 $\pm$ 0.13 & 1.45 $\pm$ 0.30 & 1.58 $\pm$ 0.13 & 3.59 $\pm$ 1.31 & 3.72 $\pm$ 1.49 \\
  FlowBatt (ours) & \textbf{0.07 $\pm$ 0.01} & \underline{0.23 $\pm$ 0.01} & \underline{0.06 $\pm$ 0.01} & \underline{0.89 $\pm$ 0.06} & \textbf{0.15 $\pm$ 0.01} \\
  \bottomrule
  \end{tabular}
\end{table}

The cycle-wise SOH results at cycle 100 are summarized in \cref{tab:soh-cyclewise}. Under this pointwise evaluation setting, both FlowBatt and BatLiNet \citep{Zhang2025} substantially outperform the BatteryML baselines \citep{Zhang2023BatteryML:AnDegradation} on all datasets, indicating that trajectory-aware representations can still provide strong performance even when evaluated only at a single cycle target. FlowBatt achieves the best performance on \texttt{MATR-1} with an RMSE of 0.07$\pm$0.01\% and on \texttt{MIX-100} with 0.15$\pm$0.01\%, while BatLiNet performs best on \texttt{MATR-2}, \texttt{HUST}, and \texttt{MIX-20}. These results show that FlowBatt remains highly competitive even under a protocol that does not explicitly exploit its full-trajectory modeling capability.

\begin{table}[h]
  \centering
  \caption{\small Extended cycle-wise SOH prediction results at cycles 200 and 300, reported as RMSE (\%) across cells.}
  \label{tab:soh-cyclewise-extended}
  \footnotesize
  \begin{tabular}{@{}llccccc@{}}
  \toprule
  Method & Cycle & \texttt{MATR-1} & \texttt{MATR-2} & \texttt{HUST} & \texttt{MIX-20} & \texttt{MIX-100} \\
  \midrule
  \multirow{2}{*}{FlowBatt (ours)}
    & 200 & 0.36 $\pm$ 0.03 & 0.73 $\pm$ 0.03 & 0.18 $\pm$ 0.03 & 0.93 $\pm$ 0.02 & 1.13 $\pm$ 0.08 \\
    & 300 & 0.69 $\pm$ 0.03 & 0.94 $\pm$ 0.04 & 0.21 $\pm$ 0.03 & 1.12 $\pm$ 0.04 & 0.82 $\pm$ 0.03 \\
  \midrule
  \multirow{2}{*}{RF (BatteryML)}
    & 200 & 1.25 $\pm$ 0.35 & 1.57 $\pm$ 0.05 & 1.17 $\pm$ 0.02 & 2.30 $\pm$ 0.00 & 3.54 $\pm$ 0.87 \\
    & 300 & 1.48 $\pm$ 0.04 & 1.65 $\pm$ 0.03 & 1.18 $\pm$ 0.03 & 2.25 $\pm$ 0.02 & 1.64 $\pm$ 0.04 \\
  \midrule
  \multirow{2}{*}{MLP (BatteryML)}
    & 200 & 1.31 $\pm$ 0.11 & 1.25 $\pm$ 0.35 & 1.52 $\pm$ 0.14 & 3.98 $\pm$ 0.88 & 1.64 $\pm$ 0.02 \\
    & 300 & 1.97 $\pm$ 0.28 & 1.19 $\pm$ 0.18 & 1.51 $\pm$ 0.14 & 3.66 $\pm$ 0.55 & 3.47 $\pm$ 0.57 \\
  \midrule
  \multirow{2}{*}{BatLiNet}
    & 200 & 0.16 $\pm$ 0.03 & 0.17 $\pm$ 0.08 & 0.85 $\pm$ 0.10 & 0.24 $\pm$ 0.01 & 0.34 $\pm$ 0.02 \\
    & 300 & 0.45 $\pm$ 0.02 & 0.34 $\pm$ 0.07 & 1.14 $\pm$ 0.17 & 0.42 $\pm$ 0.03 & 0.71 $\pm$ 0.06 \\
  \bottomrule
  \end{tabular}
\end{table}

The extended cycle-wise results in \cref{tab:soh-cyclewise-extended} show that the relative performance trends remain broadly consistent beyond the standard evaluation point at cycle 100. FlowBatt continues to clearly outperform the BatteryML baselines \citep{Zhang2023BatteryML:AnDegradation} at cycles 200 and 300 across most datasets, while BatLiNet \citep{Zhang2025} remains very strong at cycle 200 on several datasets. This extended analysis further supports the conclusion that the gains of FlowBatt are not limited to a single predefined evaluation point, but remain competitive at later stages of battery degradation as well.

\subsubsection{\texorpdfstring{Uncertainty quantification}{Uncertainty quantification}}
\label[appsec]{app:uq_metrics}

We evaluate empirical prediction intervals using coverage and mean width, as discussed by \citet{sluijterman2023evaluate}. For the FlowBatt sample-count evaluation in \cref{tab:soh-uncertainty-coverage}, reference and generated trajectories are clipped from below at 80\% SOH (90\% for \texttt{MIX-20}). Let $y_{b,i}$ and $\hat y_{b,i}^{(n)}$ denote the reference and generated SOH, respectively, in percent, for test battery $b$ at cycle $i$. The index $n=1,\ldots,N$ identifies the $N$ independent generative draws from a fixed checkpoint. The nominal 90\% empirical prediction interval is
\begin{equation}
 [L_{b,i},U_{b,i}]
=\left[
q_{0.05}\!\left(\hat y_{b,i}^{(1)},\ldots,\hat y_{b,i}^{(N)}\right),
q_{0.95}\!\left(\hat y_{b,i}^{(1)},\ldots,\hat y_{b,i}^{(N)}\right)
\right], 
\end{equation}
where $q_p$ denotes the empirical quantile at probability $p$, and $L_{b,i}$ and $U_{b,i}$ are the lower and upper bounds. Coverage is the percentage of reference values satisfying $L_{b,i}\leq y_{b,i}\leq U_{b,i}$. Mean width is the average of $U_{b,i}-L_{b,i}$, reported in SOH percentage points. Both metrics are averaged over all test battery-cycle pairs across the full evaluation horizon.

\textbf{Comparison with probabilistic baselines.}
\Cref{tab:soh-uncertainty-baselines} compares FlowBatt and DiffBatt with Quantile RF and Quantile MLP under the fixed benchmark splits. Quantile RF estimates conditional quantiles following \citet{meinshausen2006quantile}, while Quantile MLP is trained with the pinball loss. Both baselines estimate the 5th and 95th percentiles directly, without generative sampling. Quantile RF uses the baseline-subtracted discharge-capacity curve from the final early-life cycle, whereas Quantile MLP uses the complete capacity matrix.

\begin{table}[htbp]
  \centering
  \caption{\small Comparison of nominal 90\% SOH prediction intervals. Coverage is reported in percent and mean width in SOH percentage points. FlowBatt and DiffBatt use ten generated trajectories per test battery. Quantile RF and Quantile MLP results are averaged over ten training seeds.}
  \label{tab:soh-uncertainty-baselines}
  \footnotesize
  \setlength{\tabcolsep}{3.5pt}
  \begin{tabular}{@{}llccccc@{}}
  \toprule
  Method & Metric & \texttt{MATR-1} & \texttt{MATR-2} & \texttt{HUST} & \texttt{MIX-20} & \texttt{MIX-100} \\
  \midrule
  \multirow{2}{*}{FlowBatt}
  & SOH coverage & 86.0\% & 67.0\% & 65.0\% & 83.6\% & 81.7\% \\
  & SOH width & 0.30\% & 0.42\% & 1.34\% & 0.52\% & 0.64\% \\
  \midrule
  \multirow{2}{*}{DiffBatt}
  & SOH coverage & 76.6\% & 64.8\% & 56.5\% & 79.8\% & 79.1\% \\
  & SOH width & 0.25\% & 0.36\% & 2.52\% & 0.97\% & 1.37\% \\
  \midrule
  \multirow{2}{*}{Quantile RF}
  & SOH coverage & 93.0\% & 96.0\% & 67.3\% & 91.9\% & 89.4\% \\
  & SOH width & 3.06\% & 5.79\% & 6.10\% & 3.82\% & 4.07\% \\
  \midrule
  \multirow{2}{*}{Quantile MLP}
  & SOH coverage & 91.4\% & 77.2\% & 69.3\% & 95.4\% & 94.4\% \\
  & SOH width & 2.10\% & 2.64\% & 6.05\% & 4.64\% & 4.89\% \\
  \bottomrule
  \end{tabular}
\end{table}

\begin{figure}[htbp]
  \centering
  \includegraphics[width=0.8\textwidth]{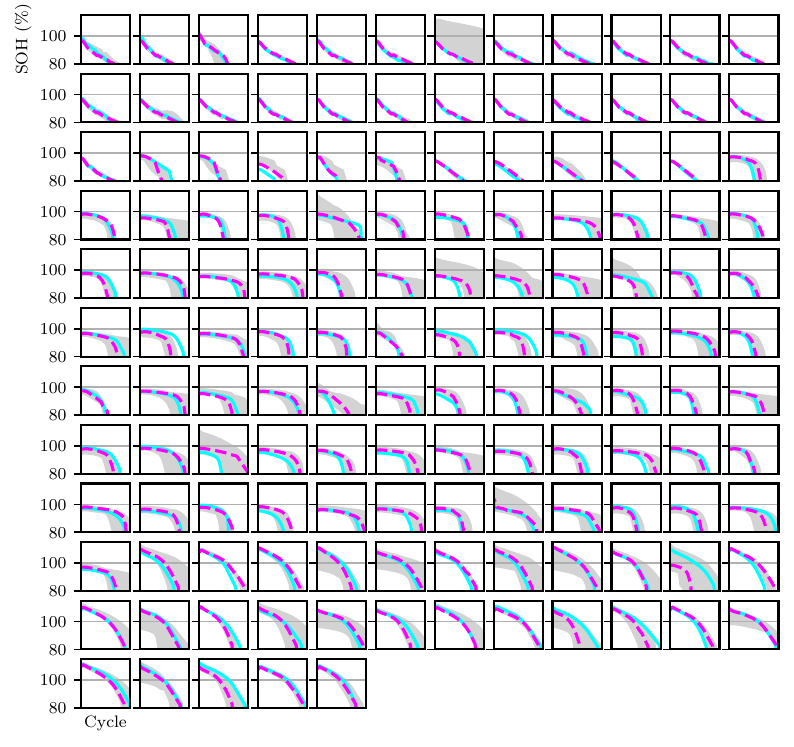}
  \caption{\small SOH predictions obtained from Quantile RF against the reference for all test samples of the \texttt{MIX-100} dataset. The pink dashed line shows the median prediction and the cyan solid line shows the reference. The gray area shows the nominal 90\% prediction interval between the 5th and 95th percentiles.}
  \label{fig:mix-quantile-rf}
\end{figure}

\begin{figure}[htbp]
  \centering
  \includegraphics[width=0.8\textwidth]{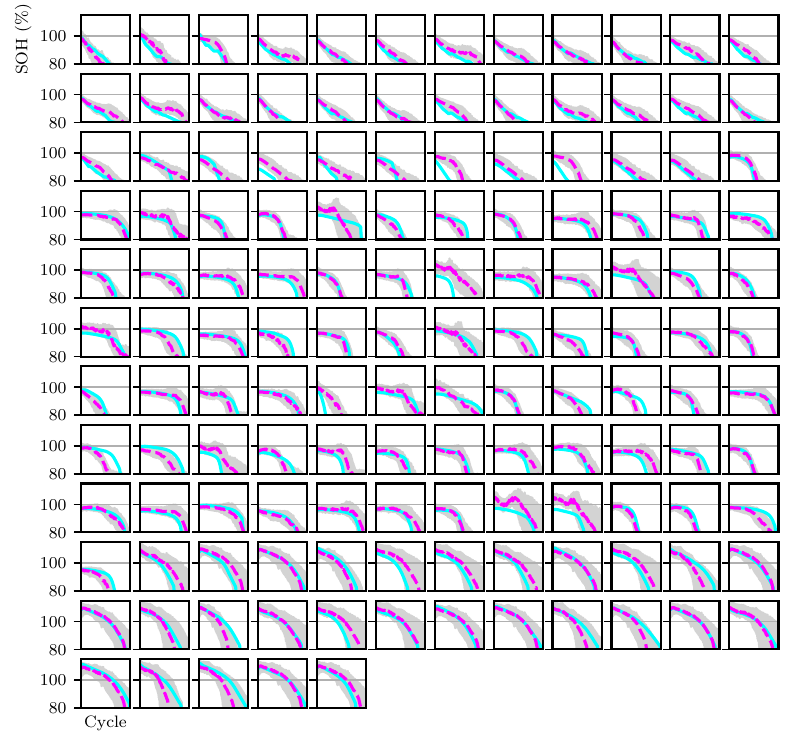}
  \caption{\small SOH predictions obtained from Quantile MLP against the reference for all test samples of the \texttt{MIX-100} dataset. The pink dashed line shows the median prediction and the cyan solid line shows the reference. The gray area shows the nominal 90\% prediction interval between the 5th and 95th percentiles.}
  \label{fig:mix-quantile-mlp}
\end{figure}

FlowBatt produces substantially narrower prediction intervals than both quantile baselines across all evaluated datasets. It also achieves higher empirical coverage than DiffBatt throughout the benchmark. On \texttt{HUST}, \texttt{MIX-20}, and \texttt{MIX-100}, this improvement is accompanied by narrower intervals, indicating a favorable coverage--width trade-off relative to DiffBatt. The quantile baselines attain higher coverage than FlowBatt, but their intervals are substantially wider. Nevertheless, both generative models remain undercovered, particularly on challenging datasets \texttt{MATR-2} and \texttt{HUST}. Despite their wider intervals, undercoverage also persists for both quantile baselines on \texttt{HUST} and for Quantile MLP on \texttt{MATR-2}. \Cref{fig:mix-quantile-rf,fig:mix-quantile-mlp} show the quantile baselines' median predictions and prediction intervals for all \texttt{MIX-100} test batteries.

\subsection{Broader impacts}
\label[appsec]{app:broader_impacts}
Our work on probabilistic battery degradation prediction has positive societal implications by supporting the transition to sustainable energy technologies. Accurate SOH forecasting and uncertainty quantification can optimize battery usage, extend lifespans, and improve the reliability of electric vehicles and stationary storage systems, thereby reducing electronic waste and resource consumption. However, there are potential negative impacts to consider. The development and training of scaled up versions of generative models, such as diffusion transformers, require significant computational resources, contributing to energy consumption and carbon emissions. Furthermore, over-reliance on data-driven predictive models without proper physical validation could lead to misestimations of battery health, potentially posing safety risks if applied blindly in critical infrastructure. We encourage future work to balance computational efficiency and tightly integrate physics-based safeguards.

\clearpage
\section*{NeurIPS Paper Checklist}

\begin{enumerate}

\item {\bf Claims}
    \item[] Question: Do the main claims made in the abstract and introduction accurately reflect the paper's contributions and scope?
    \item[] Answer: \answerYes{}
    \item[] Justification: The abstract and introduction state the main contributions of FlowBatt, namely trajectory-level SOH prediction, comparison against supervised and diffusion-based baselines, and downstream RUL evaluation. These claims are supported by the methodology and experimental results in the main paper.

\item {\bf Limitations}
    \item[] Question: Does the paper discuss the limitations of the work performed by the authors?
    \item[] Answer: \answerYes{}
    \item[] Justification: The paper discusses its limitations in \cref{sec:soh_uncertainty}, noting that uncertainty calibration remains challenging for specific datasets (\texttt{MATR-2} and \texttt{HUST}). Additionally, \cref{sec:conclusions} highlights current constraints, such as the absence of additional operational variables (e.g., temperature) and the need for stronger integration with physics-based battery aging models, framing these as areas for future work.

\item {\bf Theory assumptions and proofs}
    \item[] Question: For each theoretical result, does the paper provide the full set of assumptions and a complete (and correct) proof?
    \item[] Answer: \answerNA{}
    \item[] Justification: The paper does not present new theoretical results with theorem-proof structure. It summarizes established diffusion and flow-matching formulations as background and focuses on model design and empirical evaluation.

\item {\bf Experimental result reproducibility}
    \item[] Question: Does the paper fully disclose all the information needed to reproduce the main experimental results of the paper to the extent that it affects the main claims and/or conclusions of the paper (regardless of whether the code and data are provided or not)?
    \item[] Answer: \answerYes{}
    \item[] Justification: The paper provides comprehensive details required to reproduce the experimental results. The model architecture and preprocessing steps are described in \cref{subsec:architecture}. Furthermore, extensive details including datasets, hyperparameter settings, algorithmic steps, and baseline configurations are explicitly documented in the appendix (see \cref{app:data,app:hyperparams,app:ditblock,app:soh_BML,app:soh_BatLinet}).

\item {\bf Open access to data and code}
    \item[] Question: Does the paper provide open access to the data and code, with sufficient instructions to faithfully reproduce the main experimental results, as described in supplemental material?
    \item[] Answer: \answerYes{}
    \item[] Justification: All benchmark datasets used in this study are publicly available. The source code, pre-trained models, and instructions for reproducibility are publicly available at \url{https://github.com/HamidrezaEiv/FlowBatt}.

\item {\bf Experimental setting/details}
    \item[] Question: Does the paper specify all the training and test details (e.g., data splits, hyperparameters, how they were chosen, type of optimizer) necessary to understand the results?
    \item[] Answer: \answerYes{}
    \item[] Justification: The paper provides comprehensive details on the experimental setup. The benchmark datasets and data splits are described in the main text and \cref{app:data}. Model architecture, preprocessing steps, and specific hyperparameters are detailed in \cref{subsec:architecture}, \cref{app:hyperparams}, and \cref{tab:hyperparams}. Furthermore, detailed configurations for the baselines are provided in \cref{tab:batteryml-soh-config} and \cref{tab:batlinet-hyperparams}.

\item {\bf Experiment statistical significance}
    \item[] Question: Does the paper report error bars suitably and correctly defined or other appropriate information about the statistical significance of the experiments?
    \item[] Answer: \answerYes{}
    \item[] Justification: The main result tables (e.g., \cref{tab:soh-trajectorywise,tab:rul-results,tab:soh-cyclewise}) report the mean and standard deviation. As detailed in the main text (\cref{sec:results}) and the appendix (\cref{app:soh_fbatt,app:soh_BML}), probabilistic models are evaluated using multiple generated trajectories to quantify predictive variability, while deterministic baselines are evaluated across multiple random initializations.

\item {\bf Experiments compute resources}
    \item[] Question: For each experiment, does the paper provide sufficient information on the computer resources (type of compute workers, memory, time of execution) needed to reproduce the experiments?
    \item[] Answer: \answerYes{}
    \item[] Justification: Information regarding the computational resources, including the hardware setup, software environment, and approximate training times, is detailed in \cref{app:compute_resources}.

\item {\bf Code of ethics}
    \item[] Question: Does the research conducted in the paper conform, in every respect, with the NeurIPS Code of Ethics \url{https://neurips.cc/public/EthicsGuidelines}?
    \item[] Answer: \answerYes{}
    \item[] Justification: The work uses publicly available battery datasets, does not involve human subjects, and is presented as a scientific modeling study intended to improve battery-health prediction and reproducibility.

\item {\bf Broader impacts}
    \item[] Question: Does the paper discuss both potential positive societal impacts and negative societal impacts of the work performed?
    \item[] Answer: \answerYes{}
    \item[] Justification: A dedicated discussion regarding both the positive societal impacts (e.g., sustainability, resource efficiency) and potential negative societal impacts (e.g., computational cost, over-reliance on ML models) is provided in \cref{app:broader_impacts}.

\item {\bf Safeguards}
    \item[] Question: Does the paper describe safeguards that have been put in place for responsible release of data or models that have a high risk for misuse (e.g., pre-trained language models, image generators, or scraped datasets)?
    \item[] Answer: \answerNA{}
    \item[] Justification: The paper does not release high-risk generative assets such as language models, image generators, or scraped datasets, and the proposed model is a domain-specific battery-degradation model.

\item {\bf Licenses for existing assets}
    \item[] Question: Are the creators or original owners of assets (e.g., code, data, models), used in the paper, properly credited and are the license and terms of use explicitly mentioned and properly respected?
    \item[] Answer: \answerYes{}
    \item[] Justification: All original creators of the utilized datasets and benchmark frameworks are properly cited in the main text and \cref{app:data}. A statement confirming adherence to the open-source licenses and terms of use of these assets has been explicitly added to \cref{app:data}.

\item {\bf New assets}
    \item[] Question: Are new assets introduced in the paper well documented and is the documentation provided alongside the assets?
    \item[] Answer: \answerYes{}
    \item[] Justification: The new assets developed in this work (the FlowBatt source code and pre-trained models) are available at \url{https://github.com/HamidrezaEiv/FlowBatt}. Comprehensive documentation, including requirements and usage instructions for training and evaluation, is provided within the included README file.

\item {\bf Crowdsourcing and research with human subjects}
    \item[] Question: For crowdsourcing experiments and research with human subjects, does the paper include the full text of instructions given to participants and screenshots, if applicable, as well as details about compensation (if any)?
    \item[] Answer: \answerNA{}
    \item[] Justification: The research relies exclusively on technical, publicly available machine learning datasets for battery degradation. It does not involve any crowdsourcing or research with human subjects.

\item {\bf Institutional review board (IRB) approvals or equivalent for research with human subjects}
    \item[] Question: Does the paper describe potential risks incurred by study participants, whether such risks were disclosed to the subjects, and whether Institutional Review Board (IRB) approvals (or an equivalent approval/review based on the requirements of your country or institution) were obtained?
    \item[] Answer: \answerNA{}
    \item[] Justification: As the research does not involve human subjects, IRB approval or equivalent ethical reviews are not applicable to this work.

\item {\bf Declaration of LLM usage}
    \item[] Question: Does the paper describe the usage of LLMs if it is an important, original, or non-standard component of the core methods in this research? Note that if the LLM is used only for writing, editing, or formatting purposes and does \emph{not} impact the core methodology, scientific rigor, or originality of the research, declaration is not required.
    \item[] Answer: \answerNA{}
    \item[] Justification: Large Language Models (LLMs) were not utilized as a component of the core methodology in this research. Any usage of LLMs was strictly limited to general language editing and formatting, which does not require a declaration according to the conference guidelines.

\end{enumerate}

\end{document}